\documentclass[lettersize,journal]{IEEEtran}

\usepackage{graphicx} %插入图片的宏包
\usepackage{float} %设置图片浮动位置的宏包
\usepackage{subfigure} %插入多图时用子图显示的宏包

\usepackage[namelimits]{amsmath} %数学公式
\usepackage{amssymb}             %数学公式
\usepackage{amsfonts}            %数学字体
\usepackage{mathrsfs}            %数学花体
\usepackage[top=2cm, bottom=2cm, left=2cm, right=2cm]{geometry}
\usepackage{color}
\usepackage{algpseudocode}
\usepackage{booktabs}
\usepackage{multirow}  % 多行合并的宏包
\usepackage[ruled,linesnumbered]{algorithm2e} 
\begin{document}

\title{CL-CaGAN: Capsule Differential Adversarial Continual Learning for Cross-Domain Hyperspectral Anomaly Detection}

\author{Jianing~Wang, \IEEEmembership{Member, IEEE,}
        Siying~Guo, %~\IEEEmembership{Fellow,~OSA,}
        Zheng Hua,
        Runhu~Huang,
        Jinyu~Hu,
        and Maoguo~Gong, \IEEEmembership{Senior Member, IEEE}
%        Biao~Hou,~\IEEEmembership{Member,~IEEE,}
%        and~Licheng~Jiao, \IEEEmembership{Fellow, IEEE}
        % <-this % stops a space
%\thanks{This paper was produced by the IEEE Publication Technology Group. They are in Piscataway, NJ.}% <-this % stops a space
\thanks{Manuscript received X X, 2023; accepted X X, 2024. This work was supported in part by the National Natural Science Foundation of China under Grant 61801353.

%Manuscript received X X, 2022; accepted X X, 2022. This work was supported in part by the National Natural Science Foundation of China under Grant 61801353 and 61971273, and in part by the GHfund B under Grant 202107020822, and in part by the Project Supported by the China Postdoctoral Science Foundation funded project (No.2018M633474)..

Jianing Wang is with the Key Laboratory of Intelligent Perception and Image Understanding of Ministry of Education of China, School of Computer Science and Technology, Xidian University, Xi’an 710071, China.

Siying Guo, Zheng Hua, Runhu Huang and Jinyu Hu are with the Key Laboratory of Intelligent Perception and Image Understanding of Ministry of Education of China, School of Artificial Intelligence, Xidian University, Xi’an 710071, China. (Corresponding author: Jianing Wang and Zheng Hua, e-mail: circuitwang@163.com,  HuaZheng79@163.com)

Maoguo Gong is with the Key Laboratory of Intelligent Perception and Image Understanding of Ministry of Education, School of Electronic Engineering, Xidian University, Xi'an 710071, China.
}}

\markboth{Journal of \LaTeX\ Class Files,~Vol.~14, No.~8, August~2021}%
{Shell \MakeLowercase{\textit{et al.}}: A Sample Article Using IEEEtran.cls for IEEE Journals}


\maketitle

\begin{abstract}
Anomaly detection (AD) has attracted remarkable attention in hyperspectral image (HSI) processing fields, most existing deep learning (DL) based algorithms indicate dramatic potential for detecting anomaly samples through specific training process under current scenario. However, the limited prior information and the catastrophic forgetting problem indicate crucial challenges for existing DL structure in open scenarios cross-domain detection. In order to improve the detection performance, a novel capsule differential adversarial continual learning framework (CL-CaGAN) is proposed to elevate the cross-scenario learning performance for facilitating the real application of DL-based structure in hyperspectral anomaly detection (HAD) task. First, a modified capsule structure with adversarial learning network is constructed to estimate the background distribution for surmounting the deficiency of prior information. To mitigate the catastrophic forgetting phenomenon, clustering-based sample replay strategy and a designed extra self-distillation regularization are integrated for merging the history and future knowledge in continual AD task, while the discriminative learning ability from previous detection scenario to current scenario are retained by the elaborately designed structure with continual learning strategy. In addition, the differentiable enhancement is enforced to augment the generation performance of the training data for further stabilizing the training process with better convergence, this procedure further efficiently consolidates the reconstruction ability of background samples. To verify the effectiveness of our proposed CL-CaGAN, we conduct experiments on several real HSIs, the results indicate that the proposed CL-CaGAN demonstrates higher detection performance and continuous learning capacity for mitigating the catastrophic forgetting under cross-domain scenarios.\end{abstract}


\section{Introduction}
\IEEEPARstart{H}{yperspectral} images (HSIs) are collected by hyperspectral sensors with hundreds or even thousands of contiguous narrow spectral bands, such higher spectral resolution creates possibilities for precisely distinguishing different materials \cite{landgrebe2002hyperspectral,goetz1985imaging,2003Remote,bioucas2013hyperspectral}. Hyperspectral anomaly detection (HAD) as one of important research fields of hyperspectral information processing has been applied in many fields, including ship detection \cite{dai2015method} and mineral exploration \cite{kruse2003comparison}. The main aim of HAD task is to find pixels that are significantly different from the background in terms of spectral signatures without any prior knowledge of target information. Therefore, it is generally accepted that an anomaly deviates from the background clutter and generally covers a small area, occupying a small proportion of the image. In some cases, anomalous targets are mixed with background and appear as mixed pixels or subpixels in real-world scenes. Therefore, the limited prior knowledge of anomaly spectral and the extremely imbalanced quantity of the target and background samples brought tough challenges for anomaly detection of HSIs.

Various traditional-based and deep learning-based (DL-based) HAD methods are proposed in last decades, which presented diversity solutions for above proposed challenges.
The statistic-based methods and the representation-based methods are mainly two mainstreams for traditional HAD task. The statistic-based methods mainly estimate the probability of anomaly sample by calculating the difference between the background and the anomaly distribution. Under the hypothesis that the background obeys a multivariate Gaussian distribution, Reed-Xiaoli (RX) \cite{reed1990adaptive} is first proposed by calculating the Mahalanobis distance between test sample and approximate background mean vector to determine the anomalies \cite{mclachlan1999mahalanobis}. Thereafter, aim to better model the background distribution, KRX \cite{kwon2005kernel} is proposed to project the data into a high-dimensional feature space to characterize the background under non-Gaussian distributions. LRX \cite{2014A} utilizes a local dual-window to analyze and simulate the background. He et al. \cite{he2023recursive} propose a recursive RX with extended multi-attribute profile (RRXEMAP) algorithm that combines the extended multi-attribute profile (EMAP) and the RX algorithm, where EMAP is used to extract the spatial structure information of HSI, and the RX detector is used to remove pixels that are prone to abnormalities to purify the background. Zhang et al.\cite{zhang2022fractional} adopt a tensor reception RX algorithm based on fractional fourier transform-based tensor (FrFT) for HAD task by selecting the fractional order of FrFT by maximizing fractional Fourier entropy (FrFE). Furthermore, Chang et al. \cite{9739787} propose the assumption that both background (BKG) and anomalies can be described by the statistical properties of the first two orders (2OS) and higher orders (HOS). In \cite{9884083}, data sphering is utilized to eliminate BKG and generate a potential anomaly component through unsupervised target detection and subspace projection techniques applied to the sphered data.

In addition, spectral-spatial based constraints \cite{9467587}, spectral-spatial isolation forest \cite{9521674} and  iterative spectral-spatial HAD \cite{10049554} have been gradually explored to sufficiently utilize the spectral–spatial information. Three entropy definitions in information theory, i.e., Shannon entropy, joint entropy, and relative entropy are incorporated with density peak clustering algorithm to construct the occurrence probability of pixels for HAD\cite{9568693}. A dummy variable trick (DVT) is developed to extend constrained energy minimization (CEM) to CEM-AD, which converted a known specific target signature $d$ imposed on CEM into an unknown specific target signature \cite{9594851}. Besides, an adaptive reference-related graph embedding (ARGE) is proposed to efficaciously obtain the low-dimensional feature and improve computational efficiency \cite{10054146}. 

Apart from aforementioned algorithms, representation-based methods are also widely used in HAD. Collaborative representation-based detector (CRD) \cite{li2014collaborative} assumes that background pixels can be approximately represented by linear combinations of their spatial neighbors through reinforced $l_2$-norm minimization on the representation weight vector. Wang et al. \cite{wang2022learning} introduce a new relaxed CR detector for HAD by utilizing a new non-global dictionary while constraining the encoding vectors of different features. 
A nonnegative-constrained joint collaborative representation (NJCR) model is developed by a union dictionary consisting of background and anomaly subdictionaries \cite{9845465}. To utilize both the sparse component and the low-rank component comprehensively, a low-rank and sparse decomposition model (LSDM) with density peak guided collaborative representation (LSDDPCRD) is proposed in \cite{9349473}, where an entropy-based adaptive fusing method is designed to combine the results obtained from the low-rank matrix and the sparse component. Chang et al. \cite{9740135} present a new concept to solve the problem of anomalies being sandwiched between the background and noise during the background suppression (BS) process in HAD tasks, where the first two stages are used to solve the problem between the background and anomalies, and the sparsity cardinality (SC) is used to remove non-Gaussian noises and interferers from anomalies. Gao et al. \cite{gao2023hyperspectral} design an anomaly detection method with a chessboard topology framework (CTAD) to adaptively extract detailed information of land cover from dissected images. Zhang et al. \cite{zhang2015low} propose the HAD Mahalanobis distance method (LSMAD) based on  low-rank and sparse matrix decomposition technique. Furthermore, a low-rank sparse representation (LRASR) HAD method is proposed, in which a background dictionary is introduced and sparsity constraints are imposed on the representation coefficients \cite{2016Anomaly}. Recently, a novel enhanced total variation (ETV) with an endmember background dictionary (EBD) is designed to be used on the row vectors of the representation coefficient matrix to enhance the spatial structure of an HSI \cite{9615061}. Furthermore, a tensor-based HAD method is taken into account with prior physical constraints by applying linear TV regularization.
Furthermore, a local spatial constraint and total variation (LSC-TV) is designed based on the F-norm to force the background within the uniform spectral features, and nonisotropic TV is introduced into the proposed LSC model by using the correlation of first-order neighborhoods \cite{9512024}. However, most of above mentioned HAD algorithms are mainly based on the obtained entire image as the detection input, while it is liable to increase the difficulty in memory application and the generalization of complex modeling process.

%\textcolor{blue}{Liu et al. \cite{10194330} proposed a real-time online processing version of R-AD, referred to as RTMLMB-RAD. This method is based on multiline multiband (MLMB) data. The key feature of RTMLMB-RAD is the use of a multiline multiband correlation matrix (MLMBCM) that enables recursive updating of the R-AD detector in conjunction with MLMB data acquisition. This approach ensures real-time anomaly detection while processing the data.}

Of late, DL-based HAD methods have attracted more and more attention by virtue of the feature extracting performance without manually defining data parameters \cite{2021NAS}. 
In virtue of the capability of learning hierarchical, abstract and high-level representations, the autoencoders (AE) \cite{bati2015hyperspectral} and the generative adversarial network (GAN) \cite{goodfellow2020generative} are main commonly way to generate background samples. For AE based methods, Zhao et al. \cite{zhao2018spectral} utilized a spectral-spatial stacked AE to extract spatial–spectral feature matrices, the anomalies are detected by the Mahalanob distance obtained through low-rank and sparse matrix decomposition. Xie et al. incorporated a spectral constraint strategy into an adversarial AE to obtain better discrimination representation \cite{xie2019spectral}. In \cite{wang2022deep}, the low-rank prior and the fully convolutional AE architecture are combined to calculate the low-rank regularization loss and approximately reconstruct the background. The multi-layer AE network with skip connections is used to fully extract the rich potential features and enhance the expressive ability of the network \cite{9893839}. Liu et al. propose a dual-frequency autoencoder (DFAE) detection model in which the original HSI is transformed into high-frequency components (HFCs) and low-frequency components (LFCs) before detection \cite{9715082}. Furthermore, a background-guided deformable convolutional AE is designed with three mutually supportive parts, including encoder, decoder, and background guidance modules \cite{10322774}. In order to further suppress abnormal reconstruction, an adaptive weighted loss function and an autonomous hyperspectral AD network (Auto-AD) are designed to reconstruct the background through fully convolutional AE with skip connections as well as suppress abnormal reconstruction \cite{9382262}. 

In terms of strong representation and adversarial training capability, GAN is successfully developed to estimate the background distribution and the spectral domain feature \cite{arisoy2021gan}. Because of the high ratio of background to anomalies, the generator of GAN usually indicates better learning performance for background characteristics, while the anomaly pixels can be identified 
by a higher error value compared to background pixels. Jiang et al. \cite{jiang2020semisupervised} propose a GAN-based semi-supervised framework, in which GAN is applied to estimate the background distribution for only leveraging normal samples of training, and the model is then applied to both normal and anomalous samples to distinguish anomalies. Besides, a novel frequency-to-spectrum mapping generative adversarial network (FTSGAN) for HAD is proposed to enhance depth separable features of backgrounds and anomalies in the FrFD \cite{wang2023frequency}. GAN-based methods usually adopt convolutional neural network (CNN) \cite{li2021survey} as the main part of the generator. As a powerful alternative to CNNs, capsule network 
\cite{sabour2017dynamic} (CapsNet) is introduced to learn a more equivariant representation of images that is more robust to changes in spectral and spatial relationships of objects in HSI. Inspired by the working mechanism of human visual system, capsules are groups of locally invariant neurons that learn to recognize visual entities and output activation vectors, where the length and orientation of the activation vectors represent the estimated probability of the object and its pose parameters (relative position of samples, rotation angle, and so on), respectively.
In view of this feature representation ability, CapsNets are gradually widely explored in HSI classification task \cite{wang2021dual}. Inspired by GAN and CapsNet, Jaiswal et al. \cite{jaiswal2018capsulegan} incorporate capsules within the GAN framework and provide guidelines for designing CapsNet discriminators. A dual-channel adversarial network has been designed to generate more available training samples with contexture relation information \cite{wang2021dual}. In \cite{9057548}, Li et al. propose a novel spectral learning discriminative reconstruction (SLDR) by utilizing the spectral error map (SEM) to detect anomalies, and the spectral angle distance (SAD) is introduced to constrain the model to generate latent variables reconstruction which obeys a unit Gaussian distribution.

\begin{figure*}[htb] %H为当前位置，!htb为忽略美学标准，htbp为浮动图形
\centering %图片居中
\includegraphics[scale=0.24]{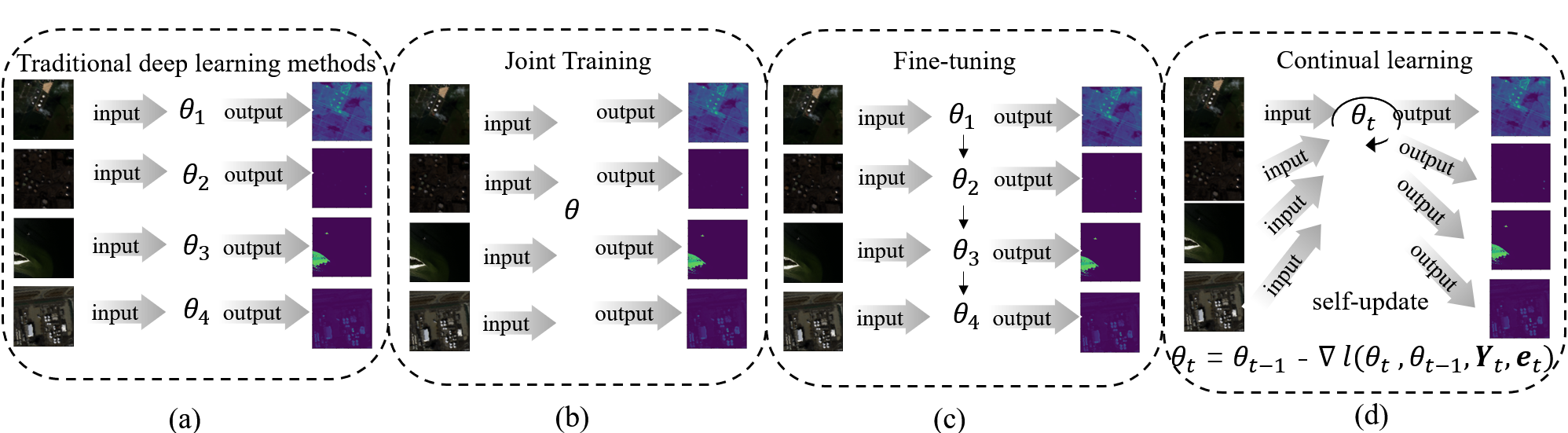} %插入图片，[]中设置图片大小，{}中是图片文件名
\caption{Comparison of different DL training model and the continuous learning method. (a) represents the traditional deep learning method, which obtains anomaly detection results by a set of independent well-trained parameters. (b) represents the joint learning method, which combines all data to train only one set of parameters for anomaly detection. (c) represents the fine-tuning method. The initialization of model parameters is based on the previous set of training parameters. (d) represents the proposed continuous learning method. The parameters of the model are continuously updated with the arrival of data, but the updated parameters will not forget the previously learned knowledge.} %最终文档中希望显示的图片标题htb
\label{compare} %用于文内引用的标签
\end{figure*}

Aforementioned existing DL-based methods mainly excel at acquiring knowledge through generalized learning behavior based on solving specific scene task from a distinct training phase. As shown in Fig.\ref{compare} (a), traditional DL-based algorithms can only deal with specific task or current scenario, which have to restart the training process when new tasks or scenarios arrive. Therefore, the specific constructed parameters of network are incapable of dealing with new tasks or scenarios thus lead to catastrophic forgetting phenomenon \cite{2013An}. To tackle this problem, joint training illustrated in Fig.\ref{compare} (b) can be regarded as multi-task optimization by parameter sharing. However, this approach requires previous training data to be available all the time, which results in an increasing demands for storage. Fine-tuning manner is shown in Fig.\ref{compare} (c), the parameters of current model is initialized from the model trained on the previous task, whereas the parameters of current model can only remember latest history knowledge. To cope with the aforementioned problems and catastrophic forgetting, continual learning (CL) \cite{parisi2019continual} (also known as lifelong learning \cite{de2021continual} and incremental learning \cite{chaudhry2018riemannian}) emerged to construct a network that can incrementally accumulate knowledge over different 
scenarios instead of retraining from scratch, therefore the network parameters can be automatically updated through customized loss functions or automatically updated exemplar set. After the training for $t$ different scenarios, the learned parameters of the model contains the ability for anomaly detection for all the previous tasks in open scenario circumstance, the procedure is briefly illustrated in Fig.\ref{compare} (d).

Numerous of CL algorithms have been proposed recently. The realization of existing CL methods can be roughly categorized into three mainstreams. 1). Replay-based methods: This kind of manner mainly preserves exemplars or synthesizes data from history knowledge to current task. Riemer et al. \cite{2018Learning} and Hou et al. \cite{hou2019learning} propose to store exemplars of old task in memory. Besides, the synthesized data are generated through generation models, which can be used to model previous knowledge for rehearsal \cite{2017FearNet,wu2018memory}. These replayed exemplars can not only be used to model the input for rehearsal but also constrained optimization of the new task loss for further preventing the interference of previous task. 2). Regularization-based methods: Regularization terms are adopted in loss function to consolidate previous knowledge of the tasks. Li et al. \cite{li2017learning} construct a distillation loss that measures the discrepancy between the output of the previous trained network and the new updated network. In \cite{zhang2020class}, a class-incremental learning paradigm with double distillation training objective is proposed to combine the two individual models trained on old classes and new classes. This line of works can avoid storing raw inputs, prioritizing privacy, and alleviating memory requirements. 3). Parameter isolation-based methods: These methods are mainly based on freezing task-specific modular parameters and growing new branches for the knowledge of the new coming tasks. The representative method such as parameter masking learns binary masks on an existing network, which can obtain a single neural network adapted to multiple tasks without affecting performance on already learned tasks \cite{2016Anomaly}. Li et al. \cite{li2019learn} employ architecture search to find the optimal structure for each sequential task. To the best of our knowledge, a great deal of research related to CL have been proposed, but by now only a few research works have been applied to the field of remote sensing for dealing with the open scenario problem \cite{bai2020class}.

Motivated by above-mentioned challenges, in this research work, we propose a novel continual capsule differentiable GAN for HAD (CL-CaGAN), in which the background is reconstructed in an adversarial manner. Specifically, in virtue of relative position and rotation angle of samples can be captured by CapsNet from a pixel vector, a modified CapsNet is incorporated into GAN (CaGAN) to enhance the position preserving and spectral discriminant ability for the reconstruction of background. Meanwhile, in order to alleviate the training instability of GAN, a differentiable data augmentation manner is exploited for all the real and pseudo samples to alleviate the training instability of GAN \cite{2020Differentiable}. Therefore, the background is reconstructed by CaGAN and the anomalies can be detected through reconstruction errors. Further for successfully applying the proposed CL-CaGAN to the open scenario circumstance, that is, the parameters of the previous task can be efficient applied to the new dataset while not forget the previous performance, we further exploit the clustering-based sample replay strategy with a designed extra distillation regularization for consolidating previous knowledge while learning new task knowledge. Therefore, the proposed CL-CaGAN can indicate more satisfying position and spectral knowledge exploitation capacity in terms of differential Capsule GAN structure. Meanwhile, the distillation-based regularization term with the clustering-based replay learning buffer also efficiently alleviates the catastrophic forgetting problem in open scenario situation. 

We then highlight the notable contributions of the proposed CL-CaGAN as follows.

• CL-CaGAN presents remarkable cross-scenario anomaly detection performance through continual learning manner by imposing (i) clustering-based replay strategy for preserving the history background and current background knowledge. (ii) an extra distillation regularization term is incorporate with differential capsule adversarial learning structure to mitigate catastrophic forgetting problem caused by cross-scenario phenomenon. To our best knowledge by now, the proposed CL-CaGAN is the first work dedicate in mitigating the catastrophic forgetting in cross-domain HAD area. 

• CL-CaGAN cooperates AE structure and a modified capsule structure in an elegant way as the generator and discriminator in GAN structure for effectively learning the representative spectral characteristics of background distribution. Therefore, the representative reconstruction of background can be more efficiently preserved in this proposed structure.

• Differentiable data augmentation strategy is incorporated into CaGAN for simultaneously augmenting real and pseudo data in generator and discriminator. This augmentation enables gradients to be efficiently propagated to the generator and discriminator, and maintains the dynamic balance of the training procedure. 

• Compared with several state-of-the-art methods via comprehensive experiments in accuracy and detection performance, the proposed CL-CaGAN presents more satisfying capability for background generation and anomaly detection. Meanwhile, because of the elaborate structure cooperation with continual learning manner, CL-CaGAN indicates more robust performance for cross-scenario detection, which paves a new way for practical application of DL structure in open scenario cross-domain anomaly detection circumstance.

The rest of this paper is organized as follows. In Section I, we mainly introduce the related developments and challenges for HAD tasks. In Section II, the details of the proposed CL-CaGAN framework is introduced. In Section III, the experimental settings and the comparison results are illustrated and discussed. Finally, the conclusions and discussions are drawn in Section IV.

\begin{figure}[htb] %H为当前位置，!htb为忽略美学标准，htbp为浮动图形
\centering %图片居中
\includegraphics[height=6.5cm,width=8.2cm]{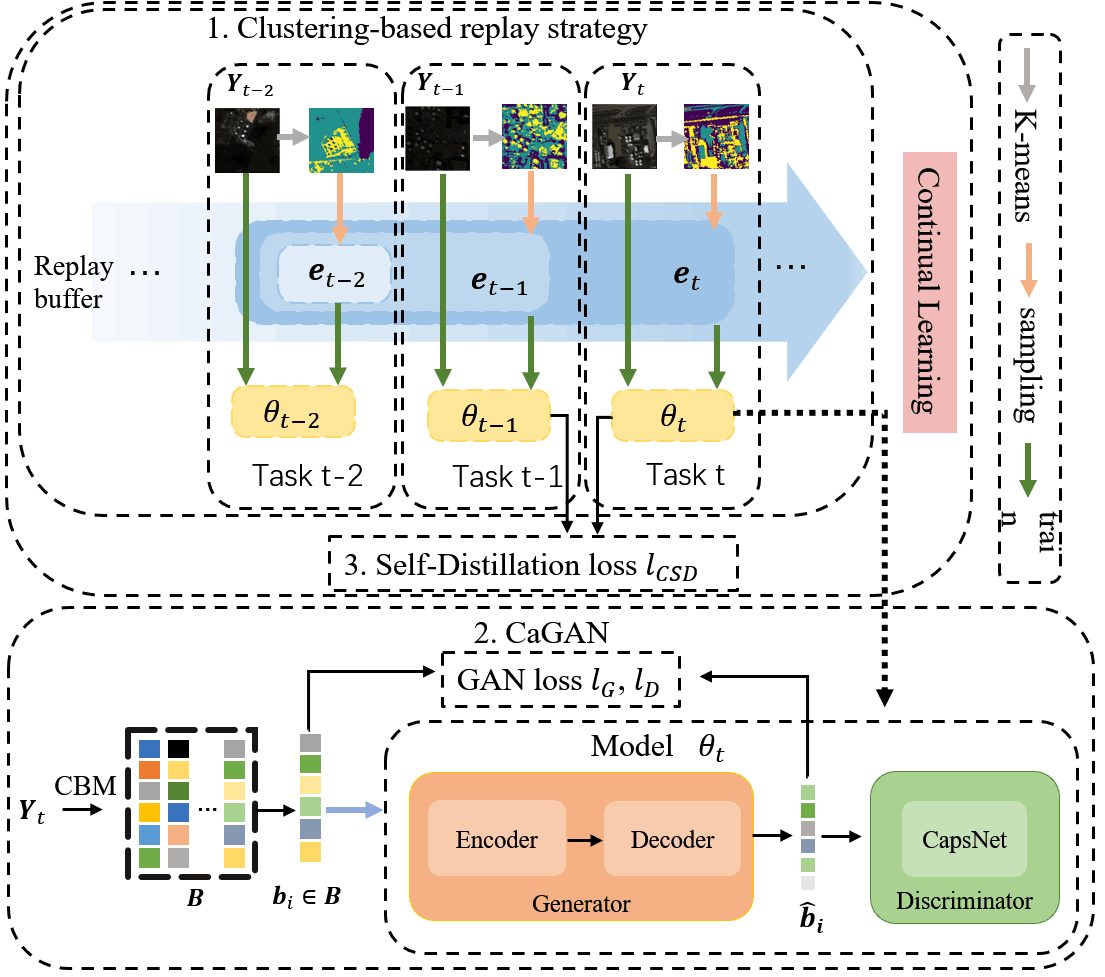}

\caption{The overview flowchart of the proposed CL-CaGAN for open scenario HAD. The entire CL-CaGAN for continuous anomaly detection process is mainly divided into three parts. 1. The cluster-based replay strategy: the representative pixels are retained in the task flow and in subsequent data to prevent forgetting phenomenon. 2. The proposed CaGAN framework: the specific structure for continuous learning AD task with cascaded generators and discriminators. 3. The proposed self-distillation loss function $L_{CSD}$ is designed to constrain the magnitude of parameter updates for preventing catastrophic forgetting.} %最终文档中希望显示的图片标题htb
\label{CL-CaGAN} %用于文内引用的标签
\end{figure}

\section{Methodology}

In this section, the proposed CL-CaGAN method for open scenario HAD is illustrated in detail. The overview flowchart of CL-CaGAN is shown in Fig.\ref{CL-CaGAN}, which mainly includes three steps: 1. Continuous exemplar replay strategy for maintaining representative background samples from previous tasks. 2. CaGAN structure for one specialized HAD task. 3. In continual learning part, the adversarial loss with continual self-distillation term is constructed to integrate historical information with current information.

\subsection{Clustering-based replay strategy}

The most challenge of CL is to cope with catastrophic problem caused by the lack of previous training data. Suppose there are $t$ different tasks with respect to datasets $\textbf{{Y}}_{1},\textbf{Y}_2,...,\textbf{Y}_t$, where $\textbf{{Y}}_{i}(i\in1,2,...,t)$ stands for the dataset coming in the $i$-th training scenario in open scenario HAD. We construct a replay buffer $\textbf{e}_t={s(\textbf{Y}_1),s(\textbf{Y}_2),...,s(\textbf{Y}_t)}$ to maintain representative background samples from previous tasks. $s$ represents the adaptive exemplar replay strategy which is utilized to select representative background samples of the input dataset. All the HAD tasks are treated equally in this procedure, that means the replay buffer could be adaptively adjusted when it encounters new tasks.

Specifically, considering the imbalanced number of exemplars from different tasks will affect the performance of CL procedure, the effect of task with fewer exemplars are apt to be degraded rapidly. To construct robust exemplar selection for all the arrived tasks, we propose an adaptive exemplar replay strategy with two objectives. 1). Select representative background samples: the representative background sample should be selected approximate to distribution of overall data. To be in line with this property, suppose there are $N_t$ samples in current task $\textbf{Y}_t$, the whole data set $\textbf{Y}_t$ is clustered to three groups for retaining the discriminative feature of current task in the memory through k-means \cite{macqueen1967some}, where each group contains $N_{i,t}(i=1,2,3)$ exemplars, respectively. 2). Construct a compromise selection strategy: the selection strategy should be adaptive to multi-task circumstance and insensitive to data distribution. Considering limited storage, we preserve $K$ representative background samples to update replay buffer $\textbf{e}_t$ for $t$-th task, where $\frac{{K\ast N}_{i,t}}{N_t}$ samples closest to each cluster center are selected as representative background samples in replay buffer. Mathematically, the proposed adaptive background sample selection strategy for the current $t$-th task can be formulated as
\begin{equation}
\begin{split}
\label{eq2}
s\left(\textbf{Y}_t\right)=\bigcup_{i=1}^{3}{\textbf{{KM}}_i\left[:\left\lfloor\frac{{K\ast N}_{i,t}}{N_t}\right\rfloor\right]}, 
\end{split}
\end{equation}
where $\textbf{{KM}}_i$ denotes the $i$-th group clustered by k-means. Note that the background samples in $\textbf{{KM}}_i$ are arrangement in ascending order of distance from the cluster center. $\ \textbf{{KM}}_i\left[:\frac{{K\ast N}_{i,t}}{N_t}\right]$ means a subset $\{\textbf{{KM}}_i\left[1\right],\textbf{{KM}}_i\left[2\right],...,\textbf{{KM}}_i\left[\left\lfloor\frac{{K\ast N}_{i,t}}{N_t}\right\rfloor\right]\}$ of $\textbf{{KM}}_i$, $\left\lfloor\bullet\right\rfloor$ is the floor operation. Furthermore, the replay buffer can be updated as
\begin{equation}
\begin{split}
\label{eq3}
\textbf{e}_t\gets\ \textbf{e}_{t-1}\cup\ s\left(\textbf{Y}_t\right),\textbf{e}_{t-1}=\phi\ when\ t=1
\end{split}
\end{equation}

By the replay strategy constructed in equation (\ref{eq3}), it is ensured that the available memory budget are maximum utilized for $K$ exemplars per task. The replay buffer $\textbf{e}_t$ is embedded with self-distillation regularization to integrate previous knowledge (self-distillation will be discuss in the following part).

%\begin{figure}[htb] %%H为当前位置，!htb为忽略美学标准，htbp为浮动图形
%\centering %图片居中

%\includegraphics[width=0.45\textwidth]{picture/fig2}
%\caption{ Illustration for the proposed CL routing.} %最终文档中希望显示的图片标题htb
%\label{CL}
%\end{figure}

\subsection{CaGAN}

\begin{figure*}[htb] %H为当前位置，!htb为忽略美学标准，htbp为浮动图形
\centering %图片居中
\includegraphics[width=0.9\textwidth]{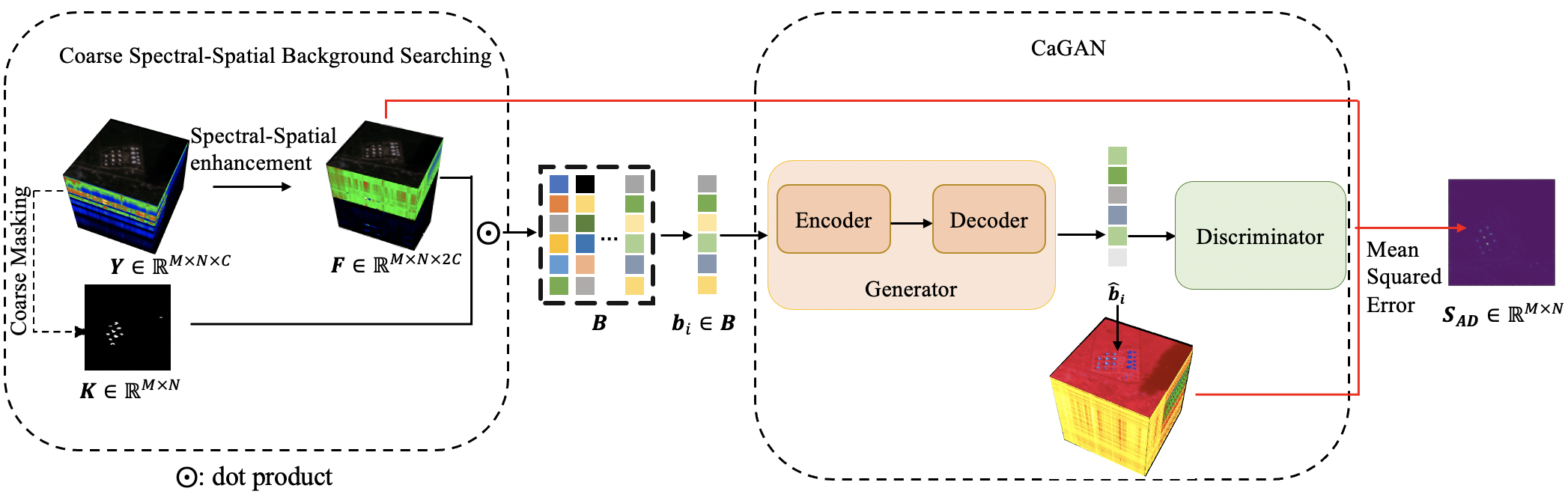} %插入图片，[]中设置图片大小，{}中是图片文件名
\caption{Overview of the proposed CaGAN for HAD. The CaGAN structure mainly contains three components.1.The coarse spectral-spatial background searching. 2.The Generator structure. 3.The Discriminator structure.} %最终文档中希望显示的图片标题htb
\label{CaGANHAD} %用于文内引用的标签
\end{figure*}

The overview of the proposed CaGAN approach for current HAD scenario task is shown in Fig.\ref{CaGANHAD}, which is mainly composed of three main components: coarse spectral-spatial background searching, and elaborately designed asymmetric generator structure and discriminator structure.
%to generate a pure training set unaffected by anomalous samples, 
\subsubsection{Coarse Spectral-Spatial Background Searching}

Mathematically, given a HSI containing $M\times N$ pixels with $C$ channels as $\textbf{Y}\in\mathbb{R}^{M\times N\times C}=\{\textbf{y}_{i,j}\in\mathbb{R}^{ C}\}_{i=1,j=1}^{i=M,j=N}$, where $\textbf{Y}=\textbf{Y}_i\left(i\in1,2,\ldots,t\right)$ is an specific HAD scene in open scenario.$\ \textbf{y}_{i,j}\in\mathbb{R}^{C}$ represents the spectral vector with the coordinate of $(i,j)$ in $\textbf{Y}$. $\textbf{Y}$ can be split into anomaly samples set $\textbf{A}$ and background samples set $\textbf{B}$, i.e.,$\textbf{Y}=\left[\textbf{A},\textbf{B}\right]$, $\textbf{A}\cup\textbf{B}=\textbf{Y}$ and $\textbf{A}\cap\textbf{B}=\emptyset$. The learning objective of the proposed CaGAN is to capture the representative feature of $\textbf{B}$ such that the distribution of reconstructed background samples ${\hat{\textbf{B}}}=G\left({\textbf{B}}\right)$ can be approximate to $\textbf{B}$.

The main challenge in HAD is to estimate and reconstruct the data distribution of background without any prior knowledge. However, the entire HSI involved as training set may be contaminated by anomalies. In order to mitigate the contamination caused by anomalous pixels, we construct coarsely background masking (CBM) matrix to obtain a relatively pure background spectral set where pixels with high probability of belonging to background are remained. Considering spectral information can provide discrimination for background and anomalies, we utilize spectral angle mapper (SAM) to measure the distance of adjacent pixels in local region to determine anomalies\cite{SAM}. Pixels with similarity value greater than the threshold are remained as background samples, otherwise, they will be rejected as anomalies. As a result, the background masking matrix $\textbf{{K}}=\{{\textbf{{k}}_{i,j}}\}_{i=1,j=1}^{i=M,j=N}$ can be generated as:
\begin{equation}
\begin{split}
\label{eq5}
SAM{\left(\textbf{y}_{i,j},\textbf{y}_{i,j+1}\right)}=\frac{\textbf{y}_{i,j}\cdot \textbf{y}_{i,j+1}}{\left|\left|\textbf{y}_{i,j}\right|\right|\times\left|\left|\textbf{y}_{i,j+1}\right|\right|}
\end{split}
\end{equation}
\begin{equation}
\begin{split}
\textbf{k}_{i,j}=\left\{
\begin{aligned}
\label{eq6}
0 & , & SAM{\left(\textbf{y}_{i,j},\textbf{y}_{i,j+1}\right)}\geq \beta, \\
1 & , & SAM{\left(\textbf{y}_{i,j},\textbf{y}_{i,j+1}\right)}< \beta.
\end{aligned}
\right.
\end{split}
\end{equation}
where $\beta$ is the threshold of background pixels’ similarity. By now, a CBM matrix is generated where background pixels are assigned with “0” and anomaly pixels with “1”.

As we know that spectral feature usually plays the key role for HAD, meanwhile, spatial correlation can enhance the representative and smoothness of characteristics. Further for involving spatial correlation in local region, a patch-wise spectral-spatial (SS) feature is involved to augment spatial information by calculating the mean vector in local spatial window. The mean vector in a $w\times w$ local region can be calculated as
\begin{equation}
\begin{split}
\label{eq7}
\bar{\textbf{y}}_{ij}=\frac{1}{w\ast w}\sum_{r=1}^{w^2}\textbf{y}_{ij}^r
\end{split}
\end{equation}
where $r$ is the index of the spectral vector in each $w\times w$ local region. The mean vector is concatenated with original spectral vector to fully enhance SS properties, which can be formulated as follows
\begin{equation}
\begin{split}
\label{eq8}
\textbf{f}_{ij}=\textbf{y}_{ij}\otimes\frac{1}{w\ast w}\sum_{r=1}^{w^2}\textbf{y}_{ij}^r=\textbf{y}_{ij}\otimes\bar{\textbf{y}}_{ij}
\end{split}
\end{equation}
%\begin{equation}
%\begin{split}
%\textbf{F}={\textbf{y}_{ij}\bigotimes\frac{1}{p\ast %q}\sum_{i=1}^{p}\sum_{j=1}^{q}\{\textbf{y}_{ij}\}}_{i=1,j%=1}^{i=M,j=N}=\{\textbf{f}_{ij}\}_{i=1,j=1}^{i=M,j=N}
%\end{split}
%\end{equation}
where $\textbf{f}_{ij}\in\mathbb{R}^{2C}$ vector represents original spectral information and local spatial relationships with the coordinate at the $(i,j)$, $\otimes$ represents concatenation operation, $\textbf{F}=\{\textbf{f}_{ij}\}_{i=1,j=1}^{i=M,j=N}$ is the SS feature matrix of HAD. Consequently, according to the coordinate index of the CBM $\textbf{K}$, the anomaly sample set $\textbf{A}$ and the background sample set $\textbf{B}$ are selected as
\begin{equation}
\begin{split}
\label{eq9}
\textbf{A}=\left\{\textbf{f}_{i,j}\middle|\ \textbf{k}_{i,j}=1\right\}=\{\textbf{a}_i\}_{i=1}^{n_a}
\end{split}
\end{equation}
\begin{equation}
\begin{split}
\label{eq10}
\textbf{B}=\left\{\textbf{f}_{i,j}\middle|\ \textbf{k}_{i,j}=0\right\}=\{\textbf{b}_i\}_{i=1}^{n_b}
\end{split}
\end{equation}
where $\textbf{a}_i\in\mathbb{R}^{2C}$ and $\textbf{b}_i\in\mathbb{R}^{2C}$ represent the $i$-th sample in $\textbf{A}$ and $\textbf{B}$, respectively, where $n_a$ and $n_b$ denote the number of samples in sets $\textbf{A}$ and $\textbf{B}$ with the constrain of $n_a+n_b=M\times N$. Particularly, samples in $\textbf{A}$ and $\textbf{B}$ are all containing original spectral information and local spatial relationships, and $\textbf{B}$ is adopted as training set for CaGAN and intends to supply relatively pure SS features.

\subsubsection{Generator structure}

\begin{figure}[htb] %H为当前位置，!htb为忽略美学标准，htbp为浮动图形
\centering %图片居中
\includegraphics[scale=0.42]{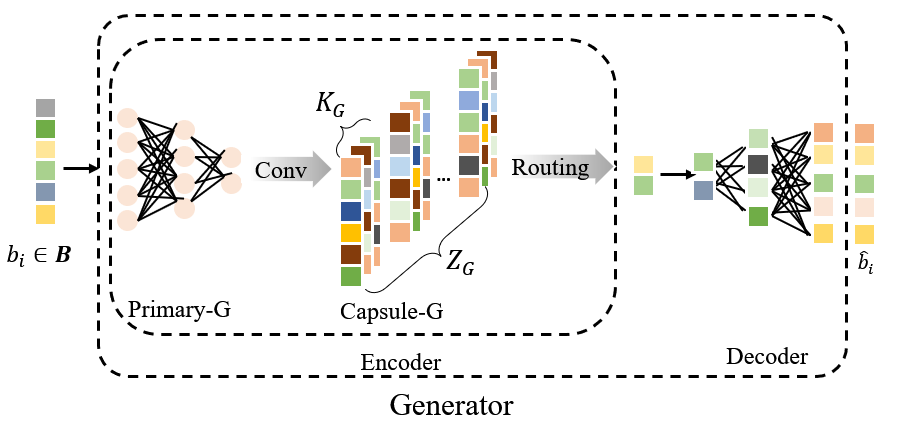} %插入图片，[]中设置图片大小，{}中是图片文件名
\caption{The Generator structure of the proposed CaGAN includes a cascaded Encoder and Decoder, where the Encoder consists of a capsule network, including two layers: Primary-G and Capsule-G. $Z_G$ in Capsule-G represents the number of capsule groups in the Generator, and $K_G$ represents the number of capsules in each group of Generator.} %最终文档中希望显示的图片标题htb
\label{Generator} %用于文内引用的标签
\end{figure}

The generator $G$ of the proposed CaGAN in HAD task is composed with AE and CapsNet for realizing background reconstruction. AE can reconstruct the input data and extract intrinsic spectral features in an end-to-end manner. A typical AE structure can be decomposed into two subnets: the encoder and the decoder. The encoder embeds input background vectors as the hidden representation in latent low-dimensional space, while the decoder reconstructs background vectors according to hidden representation. Most existing AE structures are equipped with multi-layer perceptron (MLP), which lack the spatial representation capabilities. 
Considering CNN is incompetent to accurately model the relative position between features, especially for rotated the input data. Therefore, as a variants to CNN, CapsNet is capable to efficiently enrich the representation and exploitation of existing features in virtue of vectorized feature representation ability. CapsNet can encode the rotation angle, relative position, and other instantiation parameters of features. Thereby, we construct $G$ with AE to learn hierarchical, abstract, and high-level representations of HSI, where CaspNet is incorporated to the encoder for enhancing the representative characteristics and relative relationship of features. MLP is adopted as decoder to reconstruct background samples, and the whole structure of the proposed $G$ is shown in Fig. \ref{Generator}.

There are mainly two layers in CapsNet: primary layer and capsule layer. The two layers are respectively named with Primary-G and Capsule-G in this paper. Primary-G plays a role in encoding low-level initial features. Considering MLP is specialized in extract global information of 1-dimension (1D) signals, we equipped MLP into CapsNet as primary layer, as shown in Fig.\ref{Generator}. Preliminary local-global features are obtained in Primary-G, while Capsule-G dedicates in exploiting high-level and instantiation properties. Preliminary features are arranged to $Z_G$ groups with $K_G$ capsules firstly, capsules in the same group share their weights with each other. In this way, $Z_G\times K_G$ capsules can be generated, and a $d$-dimensional vector is denoted as $\textbf{u}\in\mathbb{R}^{d}$. The orientation and length of $\textbf{u}$ represent the instantiation parameters and the probability that the input data belong to this category, respectively. Normalization is required for proper representation, the norm of $\textbf{u}$ is usually reduced by the nonlinear squashing function represented as 
\begin{equation}
\begin{split}
\label{eq11}
\overline{\textbf{u}}=\frac{\left|\left|\textbf{u}\right|\right|^2}{1+\ \left|\left|\textbf{u}\right|\right|^2}\bullet\frac{\textbf{u}}{\left|\left|\textbf{u}\right|\right|}
\end{split}
\end{equation}
where $\overline{\textbf{u}}$ denotes the normalized capsule vector. In (\ref{eq11}), the former part $\frac{\left|\left|\textbf{u}\right|\right|^2}{1+\ \left|\left|\textbf{u}\right|\right|^2}$ is to compress the norm between 0 and 1, and the latter part $\frac{\textbf{u}}{\left|\left|\textbf{u}\right|\right|}$ is to keep the orientation of the vector unchanged so that the norm of $\textbf{u}$ is compressed to 0 to 1 without changing their orientation.

After normalization, dynamic routing is utilized to connect the Capsule-G and the output. In the generator of CaGAN, the output of decoder is the reconstructed background spectral vector generated by one capsule structure. A transformation matrix $\textbf{W}_{Z_G\times K_G}$ is constructed to connect the two consecutive layers as
\begin{equation}
\begin{split}
\label{eq12}
\hat{\textbf{u}}_{Z_G\times K_G}=\textbf{W}_{Z_G\times K_G}\bullet\overline{\textbf{u}}
\end{split}
\end{equation}
where ${\hat{\textbf{u}}}_{Z_G\times K_G}$ is treated as the vote from $Z_G\times K_G$ capsules to the output capsule ${\hat{\textbf{b}}}_i\in\mathbb{R}^{2C}$. ${\hat{\textbf{b}}}_i$ is obtained by calculating a weighted sum of ${\hat{\textbf{u}}}$ attached with nonlinear squashing function, which denoted the reconstruction of $i$-th sample in background $\textbf{B}$
\begin{equation}
\begin{split}
\label{eq13}
\textbf{WS}=\sum_{j=1}^{Z_G\times K_G}{\textbf{c}_j{\hat{\textbf{u}}}_j}
\end{split}
\end{equation}
\begin{equation}
\begin{split}
\label{eq14}
{\hat{\textbf{b}}}_i=\frac{\left|\left|\textbf{WS}\right|\right|^2}{1+\ \left|\left|\textbf{WS}\right|\right|^2}\bullet\frac{\textbf{WS}}{\left|\left|\textbf{WS}\right|\right|}
\end{split}
\end{equation}
where $\textbf{WS}$ is an intermediate variable and $\textbf{c}_j$ denotes the log prior probability that the $j$-th capsule will activate the reconstructed pseudo background vectors ${\hat{\textbf{b}}}_i$. $\textbf{c}_j$ is initialized to zero and updated in each iteration as follows
\begin{equation}
\begin{split}
\label{eq15}
{\hat{\textbf{b}}}_i\bullet{\hat{\textbf{u}}}_j=\left|\left|{\hat{\textbf{b}}}_i\right|\right|\times\left|\left|{\hat{\textbf{u}}}_j\right|\right|\times cos{\left({\hat{\textbf{b}}}_i,{\hat{\textbf{u}}}_j\right)}
\end{split}
\end{equation}
\begin{equation}
\begin{split}
\label{eq16}
\textbf{c}_j\longleftarrow \textbf{c}_j+{\hat{\textbf{b}}}_i\bullet{\hat{\textbf{u}}}_j
\end{split}
\end{equation}
If the $j$-th capsule and ${\hat{\textbf{b}}}_i$ have the similar rotation and norm, which are highly correlated, will yield higher $\textbf{c}_j$. This procedure further preserves representative information of background, and realizes efficient reconstruction.

\subsubsection{Discriminator structure}

\begin{figure}[htb] %H为当前位置，!htb为忽略美学标准，htbp为浮动图形
\centering %图片居中
\includegraphics[scale=0.45]{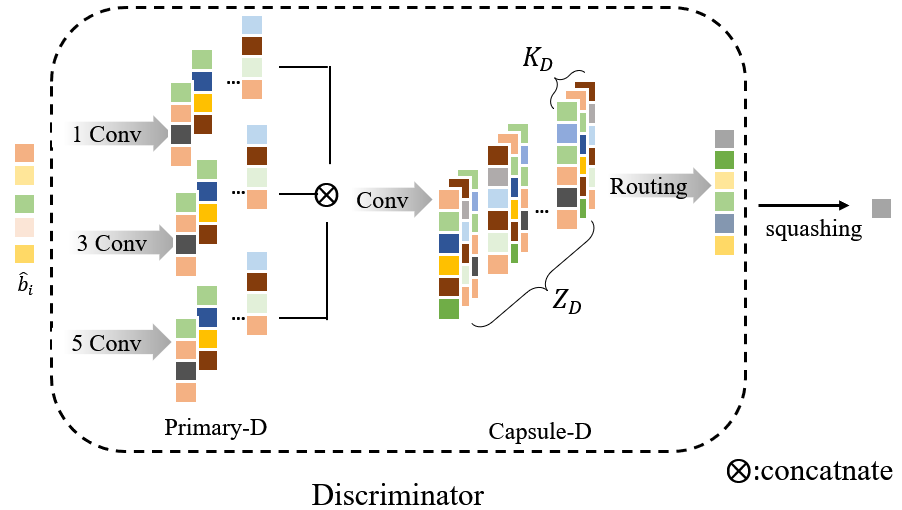} %插入图片，[]中设置图片大小，{}中是图片文件名
\caption{The discriminator structure of the proposed CaGAN. Including Primary-D and Capsule-D. Primary-D consists of convolution operations with different convolution kernel sizes. $Z_D$ in Capsule-D represents the number of groups of capsules in the Discriminator, and $K_D$ represents the number of capsules in each group of Discriminators.} %最终文档中希望显示的图片标题htb
\label{Discriminator} %用于文内引用的标签
\end{figure}

As illustrated in Fig. \ref{Discriminator}, CapsNet is exploited in discriminator to enhance discriminant spectral-spatial features. The two layers in discriminator $D$ are named as Primary-D and Capsule-D, respectively. 

To exploit the change of local–global spectral feature and present more discriminant spectral feature, multiscale convolution is constructed in Primary-D. As illustrated in Fig. \ref{Discriminator}, 1-D convolution kernels with different scales of 1, 3, and 5 is utilized to extract detailed variations. Then, the multiscale feature maps of different receptive field are concatenated in future processing. The overall change of spectral features and the change of discriminant details in a local spectral region can be efficiently exploited by multiscale convolution. The same as the Capsule-G and dynamic routing algorithm mentioned in previous part, preliminary multiscale features are arranged to capsules and the dynamic routing algorithm is utilized to calculate the output capsule vector. Specifically, the output capsule vector of discriminator is squashed by (\ref{eq11}), which is utilized to discern the real/fake of samples.

\subsection{Loss function}

The proposed adaptive exemplar replay helps to jointly and equally train the current task and retrain the previous tasks, which makes all tasks can be perceived for each other. To further consolidate knowledge of previous tasks, a continual self-distillation (CSD) loss is designed to encourage the outputs of current $t$-th network to approximate the outputs of $(t-1)$-th network.

Considering the reconstruction property of GAN, self-distillation loss can be defined as the distance of pseudo-background distribution generated between current generator and previous generator, which can be formulated as
\begin{equation}
\begin{split}
\label{eq4}
%L_{CSD}\left(\theta_t,\textbf{e}_t\right)={||G_t(\textbf{e}_t)-G_{t-1}(\textbf{e}_{t})||}_2^2
L_{CSD}={||G_t(\textbf{e}_t)-G_{t-1}(\textbf{e}_{t})||}_2^2
\end{split}
\end{equation}
where $L_{CSD}$ is the self-distillation loss term for retaining past knowledge, $\theta_t$ is the learned parameters of the $t$-th task of CaGAN structure. $G_t$ is the generator updated by $t$-th training stage and ${||\bullet||}_2^2$ denotes the $l_2$ norm which is adopted to measure distance between two distributions. 

The self-distillation loss term can obtain a slow-updated space for two adjacent training stage with the guidance of previous tasks. Otherwise, the network parameters will change uncontrollably, which is observable as catastrophic forgetting. Therefore, the proposed CaGAN structure with CSD loss can learn a more representative reconstruction background for both previous and current task through the preservation and updating in datasets and parameters of CaGAN. If the current task is not the first one, CSD loss is used in combination with the generator loss described below. The detail use of CSD loss is introduced in Algorithm 1.

During the training process, $D$ and $G$ are optimized simultaneously by the coarse background SS set $\textbf{B}$ in an adversarial manner, that is
\begin{equation}
\begin{split}
\label{eq17}
\underset{G}{\mathop{\min }}\,\underset{D}{\mathop{\max }}\,V(D,G)=&E_{\textbf{x} \sim p\left(\textbf{B}\right)}[\log\left(D\left(\textbf{x}\right)\right)+\\
&\log{\left(1-D\left(G\left(\textbf{x}\right)\right)\right)}]
\end{split}
\end{equation}

Besides, the mean squared error (MSE) between the SS background vector $\{\textbf{b}_i\}_{i=1}^{n_b}\subset \textbf{B}$ and the reconstructed pseudo-spectral vector $\{\hat{\textbf{b}_i}\}_{i=1}^{n_b}={G\left(\textbf{b}_i\right)}_{i=1}^{n_b}$ is minimized to ensure the deviation of reconstruction:
\begin{equation}
\begin{split}
\label{eq18}
L_{recon}={||{\hat{\textbf{b}}}_{i}-{\textbf{b}_i}||}_2^2
\end{split}
\end{equation}

To alleviate the training instability of GAN, we employ differentiable data augmentation function $AU$ to further augment color information (brightness, saturation, and contrast) of real and pseudo data in generator and discriminator training process of CaGAN. This procedure can efficient stabilize training, and leads to better convergence. The optimization objective function of $G$ and $D$ can be rewritten as
\begin{equation}
\begin{split}
\label{eq19}
L_G=E_{\textbf{x} \sim p\left(\textbf{B}\right)}\left[\log{\left(1-D\left(AU\left(G\left(\textbf{x}\right)\right)\right)\right)}\right]+L_{recon}
\end{split}
\end{equation}
\begin{equation}
\begin{split}
\label{eq20}
L_D=&{-E}_{\textbf{x} \sim p(\textbf{B})}[\log(D(AU(\textbf{x})))\\
&+\log{(1-D(AU(G(\textbf{x}))))}]
\end{split}
\end{equation}

During the test process, all pixels in $\textbf{F}$ are delivered into CaGAN to reconstruct pseudo-spectral vector $\hat{\textbf{F}}$. The final detection map $\textbf{S}_{AD}$ is constructed via
\begin{equation}
\begin{split}
\label{eq21}
\textbf{S}_{AD}={||\textbf{F}-\hat{\textbf{F}}||}_2^2,\ \hat{\textbf{F}}=CaGAN\left(\textbf{F}\right)
%\ \textbf{F}\in\mathbb{R}^{M\times N\times2C}
\end{split}
\end{equation}

\subsection{Summation for CL-CaGAN}

\begin{algorithm*}[!ht]
	\caption{Procedure of the Proposed CL-CaGAN Method} 
	\label{A1} 
	\KwIn{The stream HSIs $\textbf{Y}_1$,$\textbf{Y}_2$,$\ldots$,$\textbf{Y}_t$ for $t$ tasks, the threshold $\tau$=0.99 for the generation of CBM matrix, the allowed number of exemplars $K$ for each preserving task.\\}

	\textbf{Initialization: } Initialize replay buffer $\textbf{e}_1$=$\phi$, the random weight and biases of parameterized $\theta_1$ for CL-CaGAN.

	\For{ $r$=1 to $t$ tasks }
	{
		$\textbf{Y}_r\in\mathbb{R}^{M\times N\times C}={\{\boldsymbol{y}_{i,j}\in\mathbb{R}^{C}\}}_{i=1,j=1}^{i=M,j=N}$;
		
		\For{ $E$ training iterations}
		{
			generate $\textbf{k}_{i,j}$ vector in CBM by (\ref{eq5}) and (\ref{eq6})

            concact SS information with original spectral to construct $\textbf{F}=\{\textbf{f}_{ij}\}_{i=1,j=1}^{i=M,j=N}$ through (\ref{eq7}) and (\ref{eq8})
            
            construct background samples set $\textbf{B}=\left\{\textbf{f}_{i,j}\middle|\textbf{k}_{i,j}=0\right\}={\textbf{b}_i}_{i=1}^{n_b}$ by (\ref{eq10});
            
            \If  {$r$ = 1} {update $\theta_1$ by minimizing $L_G$ and $L_D$ through equation (\ref{eq19}) and (\ref{eq20});}
            
            \Else {update $\theta_r$ by minimizing $L_G+L_{CSD}$ and $L_D$ through equation (\ref{eq4}), (\ref{eq19}) and (\ref{eq20});}
            
            apply k-means to $\textbf{B}$ and obtain $N_i\left(i=1,2,3\right)$ three groups containing  exemplars respectively;
            
            The subset of $\textbf{B}$ obtained by $s\left(\textbf{B}\right)=\bigcup_{i=1}^{3}{{\textbf{KM}}_i\left[:\left\lfloor\frac{{K\ast N}_i}{n_b}\right\rfloor\right]}$, where ${\textbf{KM}}_i$ denotes the $i$-th group clustered by k-means;
            
            \If  {$r$ = 1} {update replay buffer by $\textbf{e}_1\ \gets\ s\left(\textbf{B}\right)$ ;}
            
            \Else {update replay buffer by $\textbf{e}_r\ \gets \textbf{e}_{r-1}\cup\ s\left(\textbf{B}\right)$;}
		}
	}
	Construct the detection for all tasks through equation (\ref{eq21});

	\KwOut{parameters $\theta_t$ and detection map $\textbf{S}_{AD}$ for $t$ task}
\end{algorithm*}

In this paper, a novel CaGAN structure is designed for HAD, meanwhile, the cluster-based sample replay and self-distillation loss are incorporated with CaGAN to achieve more robust performance for unending continual learning scenarios. Therefore, after training CL-CaGAN for unending $t$ HAD tasks, the well-optimized parameters is capable to detect anomalies in all $t$ tasks and obtain more stable and satisfying results for cross-scene anomaly detection. In addition, the elaborate designed CaGAN combines GAN and AE with two asymmetric CapsNet for better realizing reconstruction of background samples located around the boundary of anomalies. The whole procedure of the proposed CL-CaGAN is summarized in Algorithm \ref{A1}. 

\section{Experiments}

We evaluate the proposed method on five different real HSIs for anomaly detection \cite{kang2017hyperspectral}, which are captured by the Airborne Visible/Infrared Imaging Spectrometer (AVIRIS) sensor under different scenarios. The detail information of these datasets are illustrated in Table \ref{DataSet}.

In the following part, we present and discuss the performance of our proposed CL-CaGAN by specified HAD (part A) and CL-based open scenario HAD (part B).  
\begin{table}[!ht]
	\centering
	\caption{Details Of The Anomaly Detection Data Set}
	\label{DataSet}
	\setlength{\tabcolsep}{0.5mm}{
	\begin{tabular}{cccccc}
		\toprule  % 顶部线
		HSIs & Spatial size &	channels	& resolution/m & bands/nm &	Sensor \\
		%\midrule  % 中部线
		%Texas Coast &	100$\times$100	& 207 &	17.2 &	450 - 1350 &	AVIRIS \\
		\midrule  % 中部线
		Los Angeles-1 &	100$\times$100 &	205 &	7.1	& 430 - 860 &	AVIRIS \\
		\midrule  % 中部线
		Los Angeles-2 &	100$\times$100 &	205 &	7.1 &	430 - 860 &	AVIRIS \\
		\midrule  % 中部线
		Cat Island &	150$\times$150 &	188 &	17.2 &	400–2500 &	AVIRIS \\
		\midrule  % 中部线
		San Diego &	100$\times$100 &	193 &	7.5 &	400–2500 &	AVIRIS \\
		\midrule  % 中部线
		Bay Champagne &	100$\times$100 &	188	& 4.4 &	400–2500 &	AVIRIS \\
		%\midrule  % 中部线
		%Pavia &	150$\times$150 &	102 &	1.3 &	430 - 860 &	ROSIS \\
		%\midrule  % 中部线
		%HYDICE &	 80$\times$100 &	  175  &	 1.0 &	 400-2500 & HYDICE \\
		\bottomrule  % 底部线
	\end{tabular}
	}
	
\end{table}

\begin{figure*}[htb] %H为当前位置，!htb为忽略美学标准，htbp为浮动图形
\centering %图片居中

\subfigure[]{
\label{pseudoColor}
\begin{minipage}[t]{0.065 \linewidth}
\centering
%\includegraphics[width=13mm,height=13mm]{picture/pseudoColor/AU2.png}
%\vspace{-0.3cm}

%\includegraphics[width=13mm,height=13mm]{picture/pseudoColor/AU3.png}
%\vspace{-0.3cm}

\includegraphics[width=13mm,height=13mm]{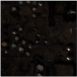}
\vspace{-0.3cm}

\includegraphics[width=13mm,height=13mm]{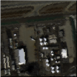}
\vspace{-0.3cm}

\includegraphics[width=13mm,height=13mm]{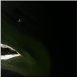}
\vspace{-0.3cm}

\includegraphics[width=13mm,height=13mm]{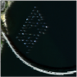}
\vspace{-0.3cm}

\includegraphics[width=13mm,height=13mm]{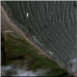}
\vspace{-0.3cm}

%\includegraphics[width=13mm,height=13mm]{picture/pseudoColor/AB4.png}
%\vspace{-0.3cm}

%\includegraphics[width=13mm,height=13mm]{picture/pseudoColor/AA4.png}
%\vspace{-0.3cm}

%\includegraphics[width=13mm,height=10.4mm]{picture/pseudoColor/HYDICE.png}
%\vspace{-0.3cm}
\end{minipage}
}%插入图片，[]中设置图片大小，{}中是图片文件名
\subfigure[]{
\label{gt}
\begin{minipage}[t]{0.065\linewidth}
\centering
%\includegraphics[width=13mm,height=13mm]{picture/gt/AU2.png}
%\vspace{-0.3cm}

%\includegraphics[width=13mm,height=13mm]{picture/gt/AU3.jpg}
%\vspace{-0.3cm}

\includegraphics[width=13mm,height=13mm]{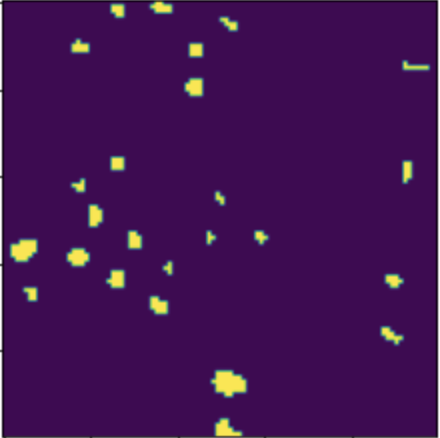}
\vspace{-0.3cm}

\includegraphics[width=13mm,height=13mm]{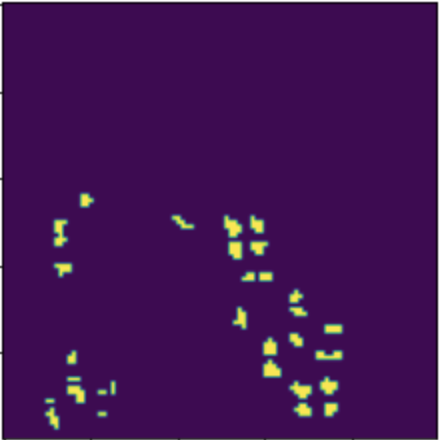}
\vspace{-0.3cm}

\includegraphics[width=13mm,height=13mm]{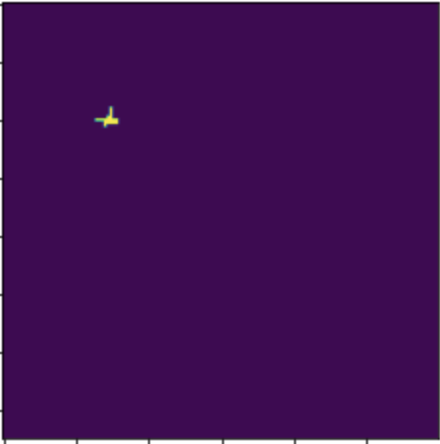}
\vspace{-0.3cm}

\includegraphics[width=13mm,height=13mm]{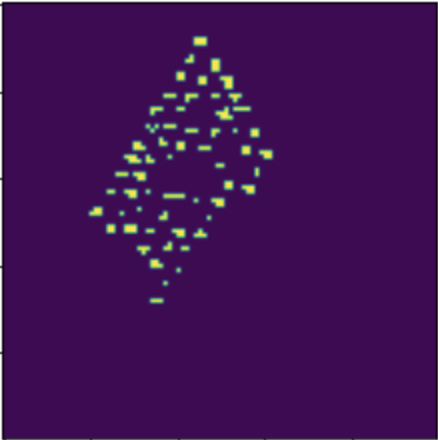}
\vspace{-0.3cm}

\includegraphics[width=13mm,height=13mm]{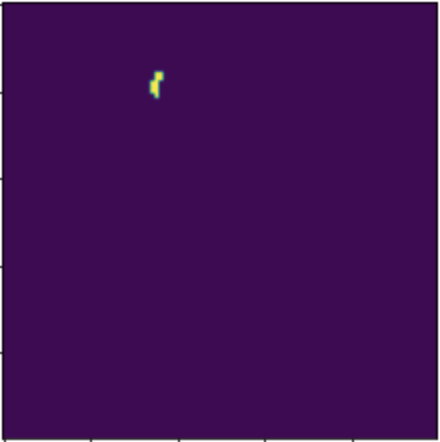}
\vspace{-0.3cm}

%\includegraphics[width=13mm,height=13mm]{picture/gt/AB4.png}
%\vspace{-0.3cm}

%\includegraphics[width=13mm,height=13mm]{picture/gt/AA4.png}
%\vspace{-0.3cm}

%\includegraphics[width=13mm,height=10.4mm]{picture/gt/gt_HYDICE.png}
%\vspace{-0.3cm}
\end{minipage}
}%插入图片，[]中设置图片大小，{}中是图片文件名\
\subfigure[]{
\label{RX}
\begin{minipage}[t]{0.065\linewidth}
\centering
%\includegraphics[width=13mm,height=13mm]{picture/RX/AU2.png}
%\vspace{-0.3cm}

%\includegraphics[width=13mm,height=13mm]{picture/RX/AU3.png}
%%\vspace{-0.3cm}

\includegraphics[width=13mm,height=13mm]{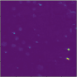}
\vspace{-0.3cm}

\includegraphics[width=13mm,height=13mm]{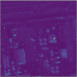}
\vspace{-0.3cm}

\includegraphics[width=13mm,height=13mm]{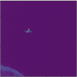}
\vspace{-0.3cm}

\includegraphics[width=13mm,height=13mm]{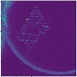}
\vspace{-0.3cm}

\includegraphics[width=13mm,height=13mm]{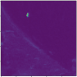}
\vspace{-0.3cm}

%\includegraphics[width=13mm,height=13mm]{picture/RX/AB4.png}
%\vspace{-0.3cm}

%\includegraphics[width=13mm,height=13mm]{picture/RX/AA4.png}
%\vspace{-0.3cm}

%\includegraphics[width=13mm,height=10.4mm]{picture/RX/RX_HYDICE.png}
%\vspace{-0.3cm}
\end{minipage}
}%插入图片，[]中设置图片大小，{}中是图片文件名\
\subfigure[]{
\label{KRX}
\begin{minipage}[t]{0.065\linewidth}
\centering
%\includegraphics[width=13mm,height=13mm]{picture/LRX/LRX_AU2.png}
%\vspace{-0.3cm}

%\includegraphics[width=13mm,height=13mm]{picture/KRX/AU3.png}
%\vspace{-0.3cm}

\includegraphics[width=13mm,height=13mm]{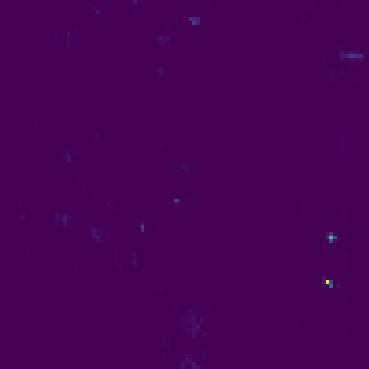}
\vspace{-0.3cm}

\includegraphics[width=13mm,height=13mm]{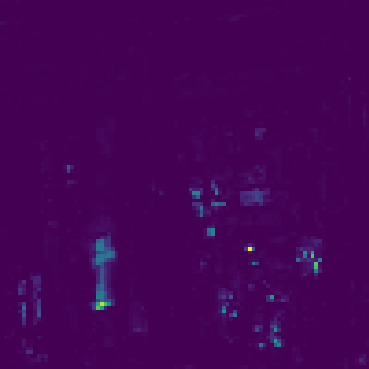}
\vspace{-0.3cm}

\includegraphics[width=13mm,height=13mm]{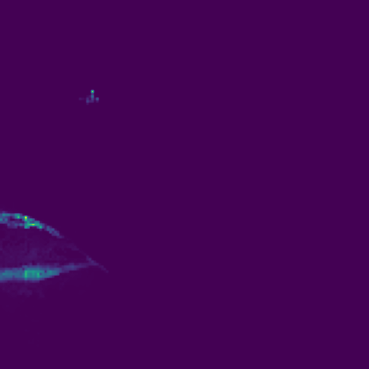}
\vspace{-0.3cm}

\includegraphics[width=13mm,height=13mm]{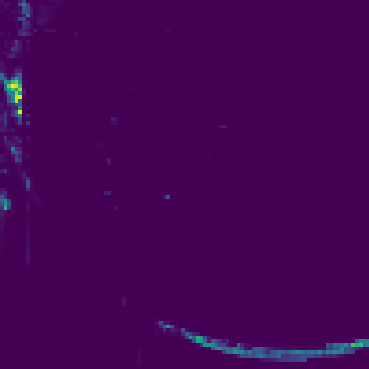}
\vspace{-0.3cm}

\includegraphics[width=13mm,height=13mm]{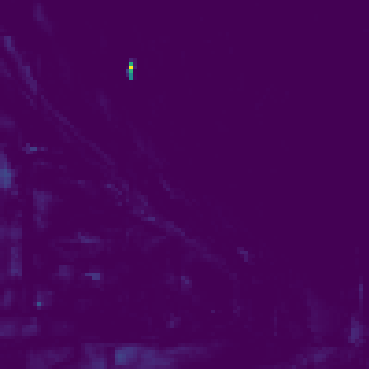}
\vspace{-0.3cm}

%\includegraphics[width=13mm,height=13mm]{picture/LRX/LRX_AB4.png}
%\vspace{-0.3cm}

%\includegraphics[width=13mm,height=13mm]{picture/KRX/AA4.png}
%\vspace{-0.3cm}

%\includegraphics[width=13mm,height=10.4mm]{picture/LRX/LRX_HYDICE.png}
%\vspace{-0.3cm}
\end{minipage}
}%插入图片，[]中设置图片大小，{}中是图片文件名\
\subfigure[]{
\label{LRX}
\begin{minipage}[t]{0.065\linewidth}
\centering
%\includegraphics[width=13mm,height=13mm]{picture/PAB-DC/AU2.png}
%\vspace{-0.3cm}

%\includegraphics[width=13mm,height=13mm]{picture/LRX/AU3.png}
%\vspace{-0.3cm}

\includegraphics[width=13mm,height=13mm]{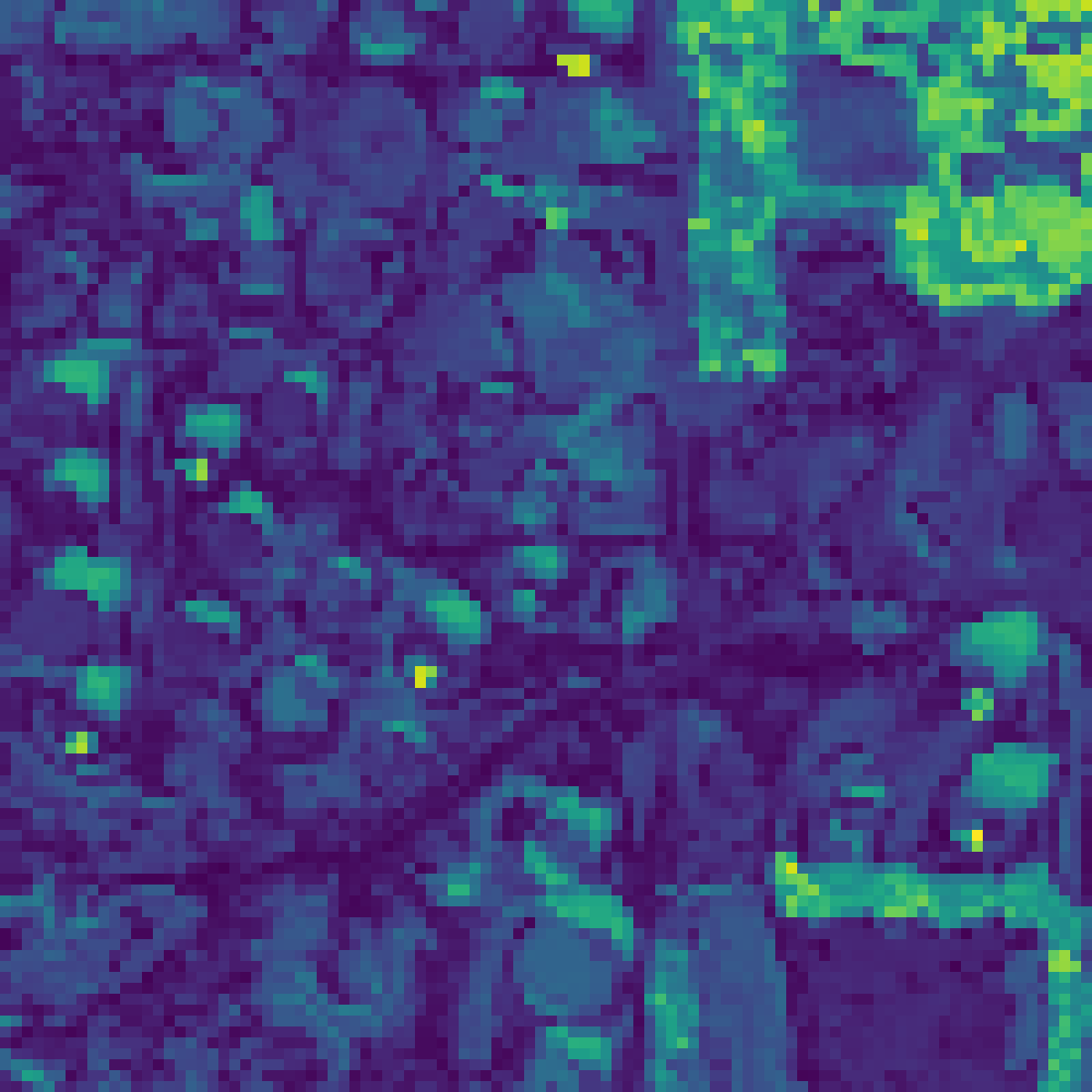}
\vspace{-0.3cm}

\includegraphics[width=13mm,height=13mm]{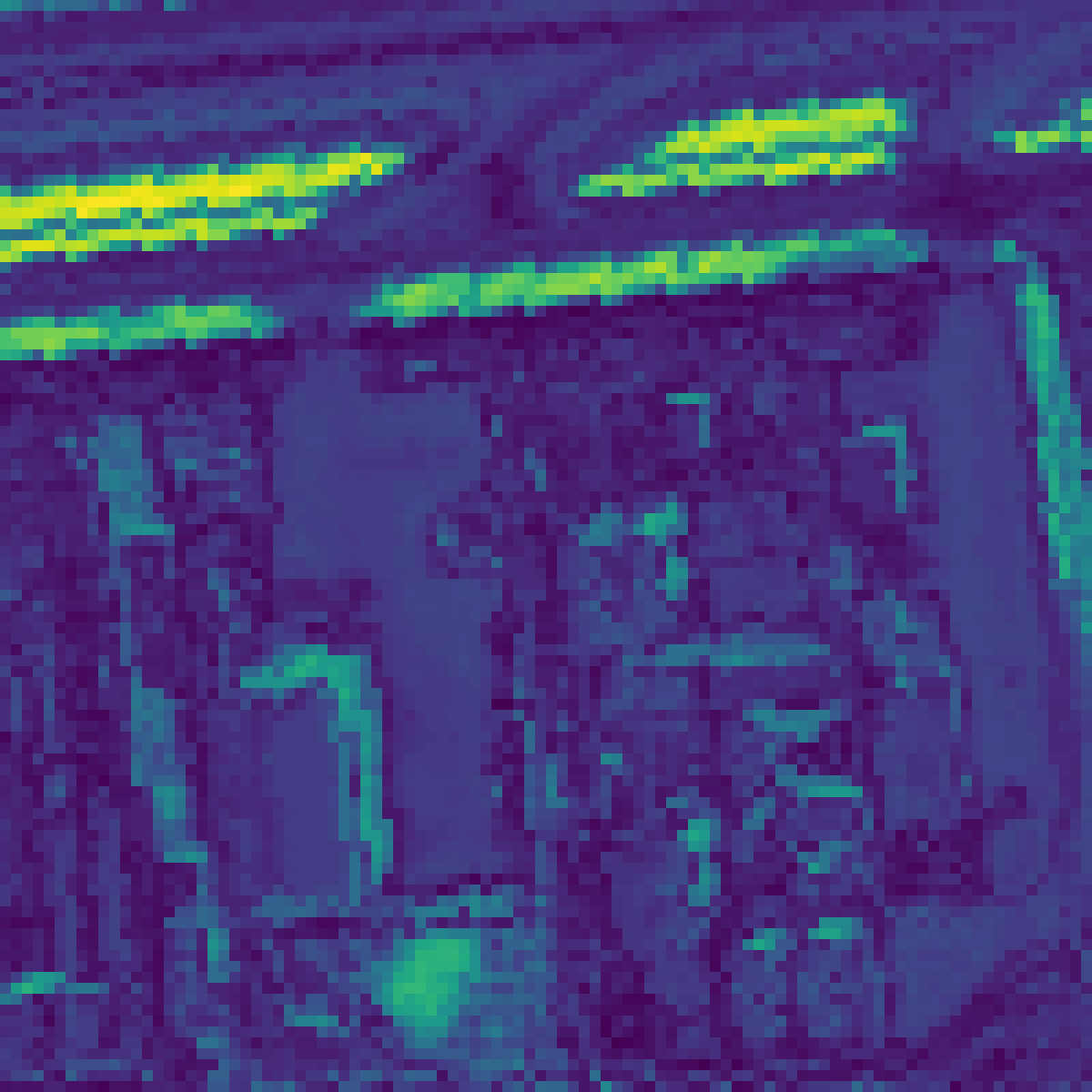}
\vspace{-0.3cm}

\includegraphics[width=13mm,height=13mm]{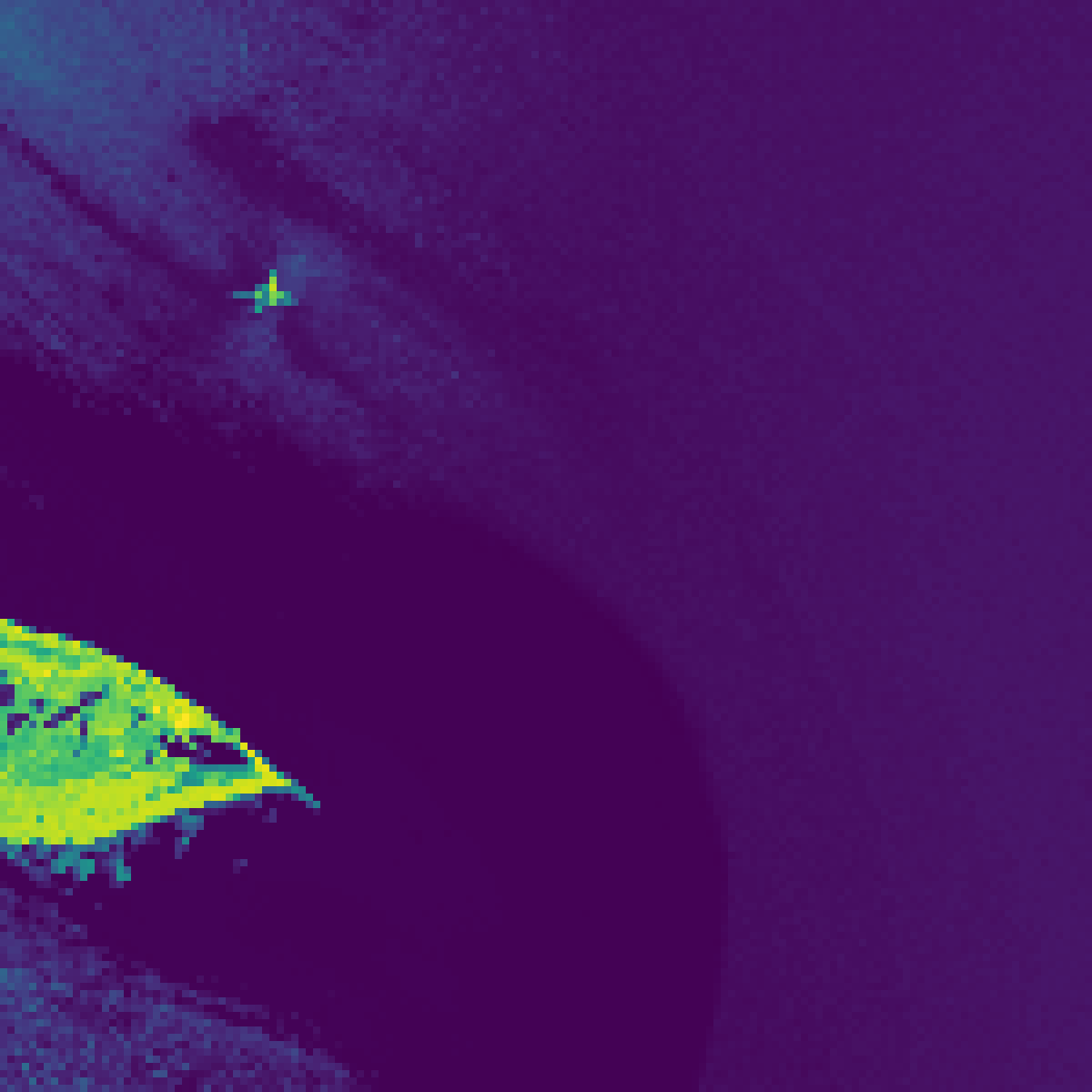}
\vspace{-0.3cm}

\includegraphics[width=13mm,height=13mm]{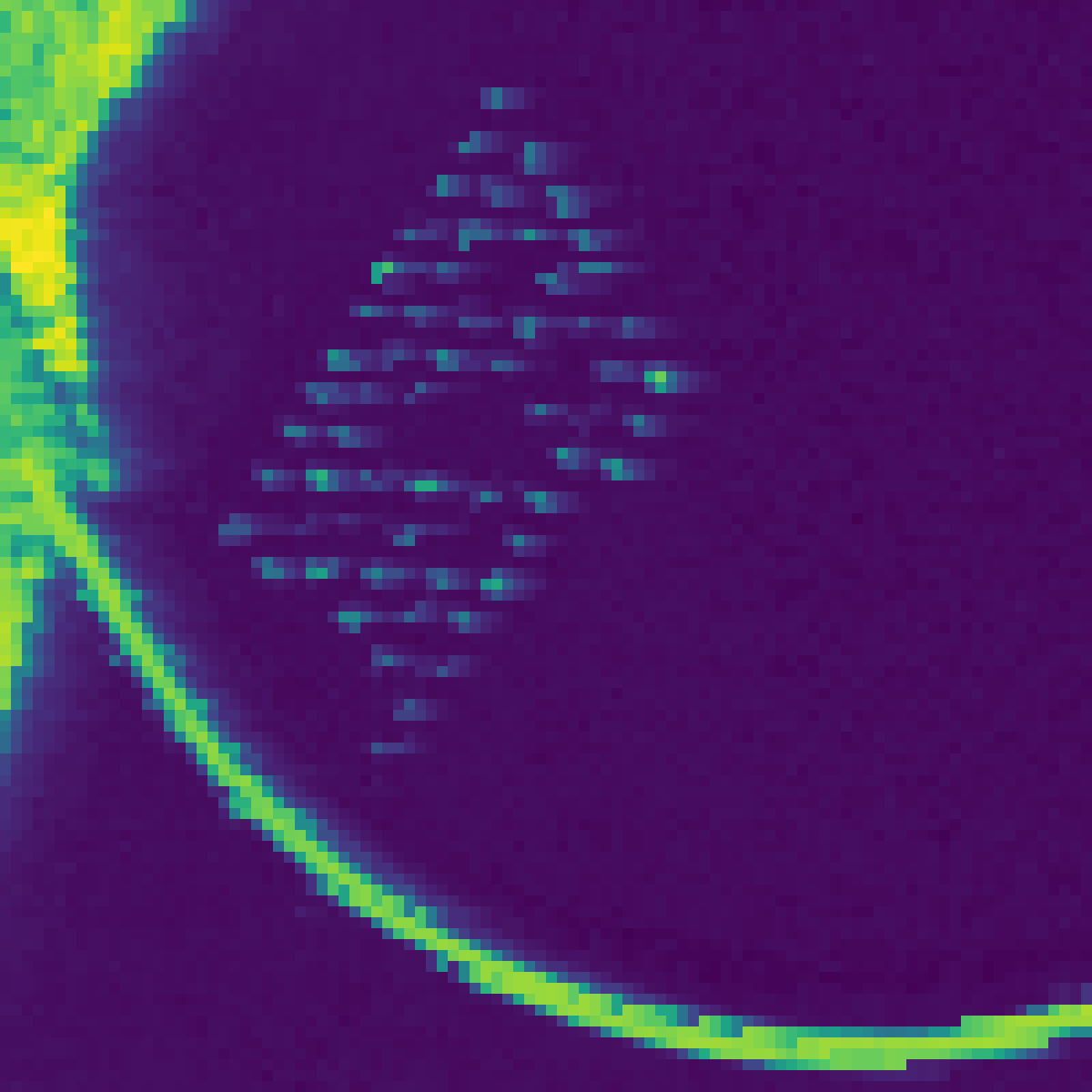}
\vspace{-0.3cm}

\includegraphics[width=13mm,height=13mm]{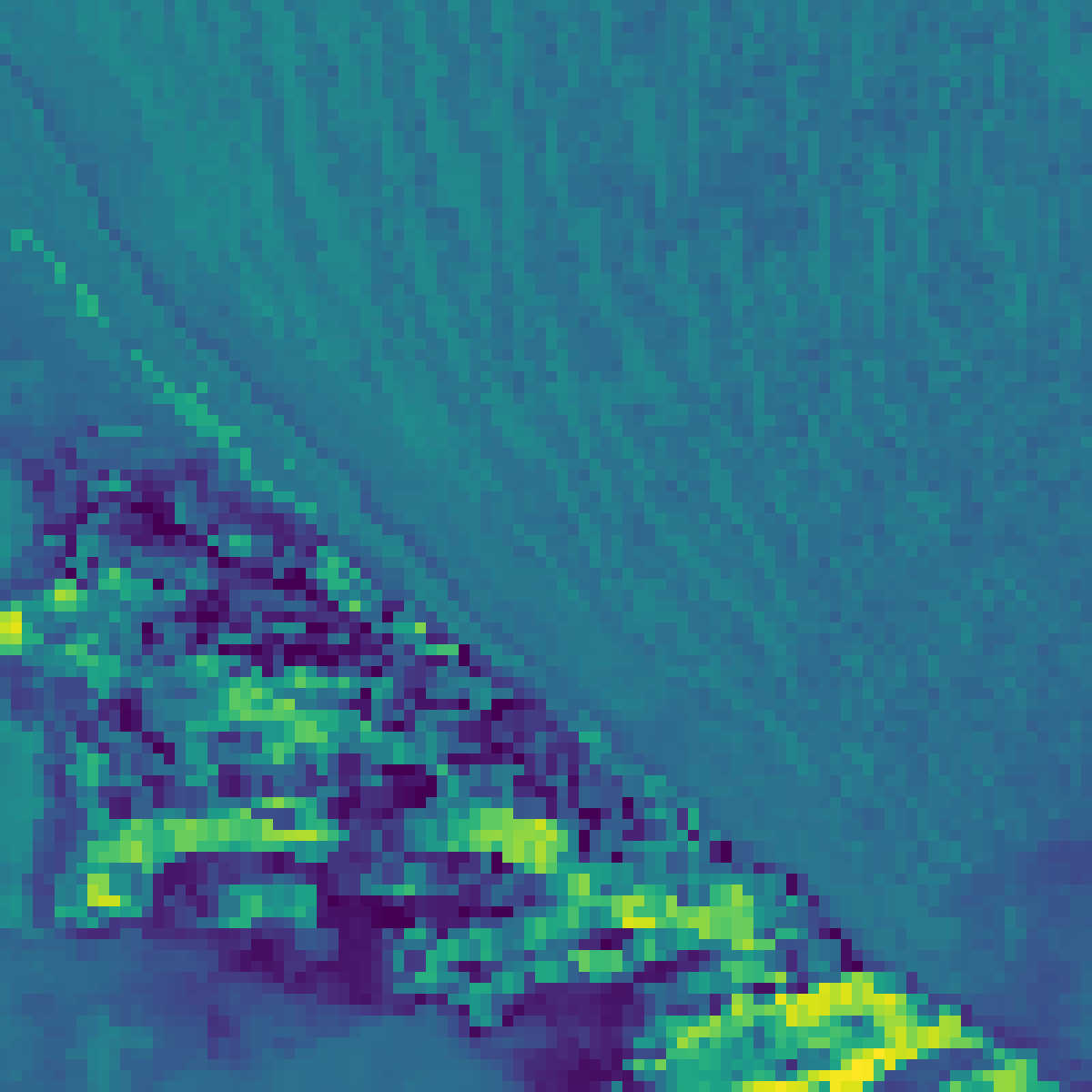}
\vspace{-0.3cm}

%\includegraphics[width=13mm,height=13mm]{picture/PAB-DC/AB4.png}
%\vspace{-0.3cm}

%\includegraphics[width=13mm,height=13mm]{picture/LRX/AA4.png}
%\vspace{-0.3cm}

%\includegraphics[width=13mm,height=10.4mm]{picture/PAB-DC/PAB_HYDICE.png}
%\vspace{-0.3cm}
\end{minipage}
}%插入图片，[]中设置图片大小，{}中是图片文件名\
\subfigure[]{
\label{CRD}
\begin{minipage}[t]{0.065\linewidth}
\centering
%\includegraphics[width=13mm,height=13mm]{picture/AED/AU2.png}
%\vspace{-0.3cm}

%\includegraphics[width=13mm,height=13mm]{picture/CRD/AU3.png}
%\vspace{-0.3cm}

\includegraphics[width=13mm,height=13mm]{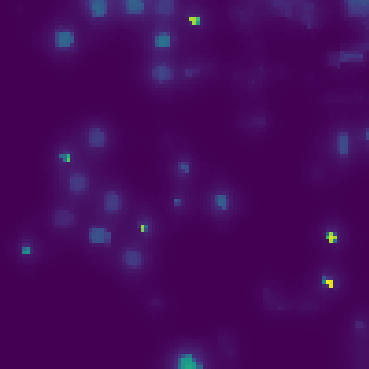}
\vspace{-0.3cm}

\includegraphics[width=13mm,height=13mm]{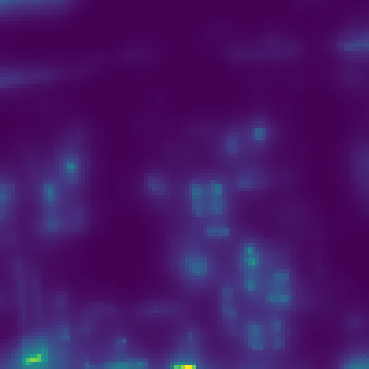}
\vspace{-0.3cm}

\includegraphics[width=13mm,height=13mm]{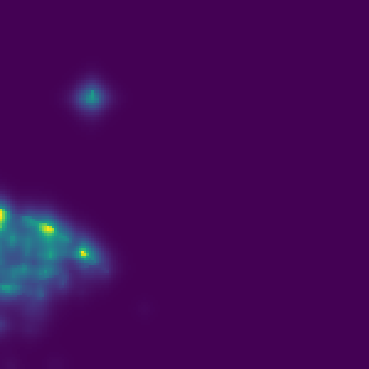}
\vspace{-0.3cm}

\includegraphics[width=13mm,height=13mm]{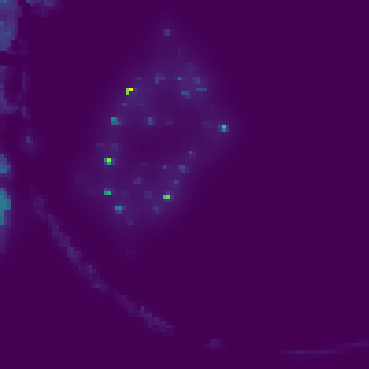}
\vspace{-0.3cm}

\includegraphics[width=13mm,height=13mm]{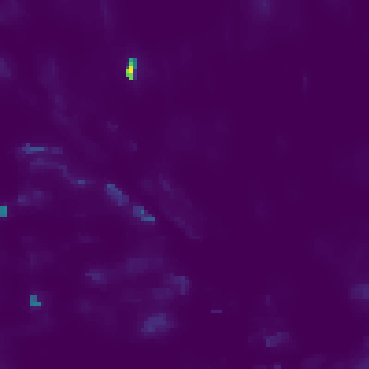}
\vspace{-0.3cm}

%\includegraphics[width=13mm,height=13mm]{picture/AED/AB4.png}
%\vspace{-0.3cm}

%\includegraphics[width=13mm,height=13mm]{picture/CRD/AA4.png}
%\vspace{-0.3cm}

%\includegraphics[width=13mm,height=10.4mm]{picture/AED/AED_HYDICE.png}
%\vspace{-0.3cm}
\end{minipage}
}%插入图片，[]中设置图片大小，{}中是图片文件名\
\subfigure[]{
\label{AED}
\begin{minipage}[t]{0.065\linewidth}
\centering
%\includegraphics[width=13mm,height=13mm]{picture/EAS/EAS_RX_AU2.png}
%\vspace{-0.3cm}

%\includegraphics[width=13mm,height=13mm]{picture/AED/AU3.png}
%\vspace{-0.3cm}

\includegraphics[width=13mm,height=13mm]{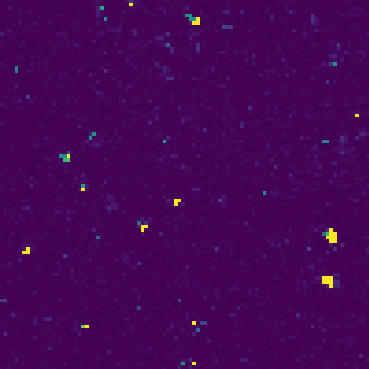}
\vspace{-0.3cm}

\includegraphics[width=13mm,height=13mm]{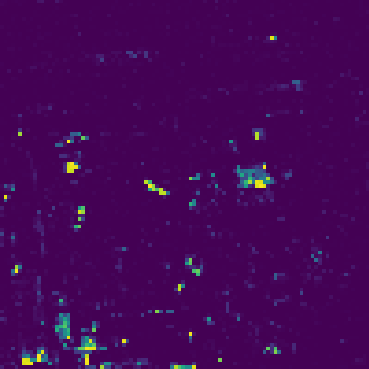}
\vspace{-0.3cm}

\includegraphics[width=13mm,height=13mm]{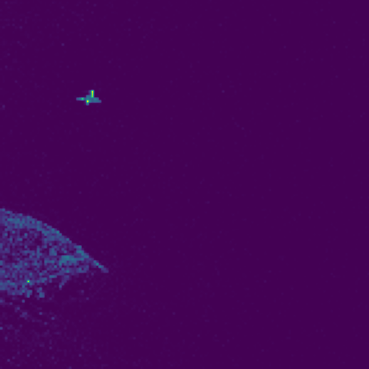}
\vspace{-0.3cm}

\includegraphics[width=13mm,height=13mm]{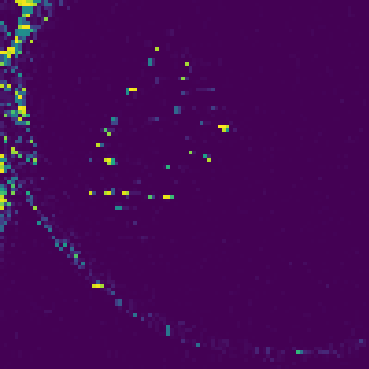}
\vspace{-0.3cm}

\includegraphics[width=13mm,height=13mm]{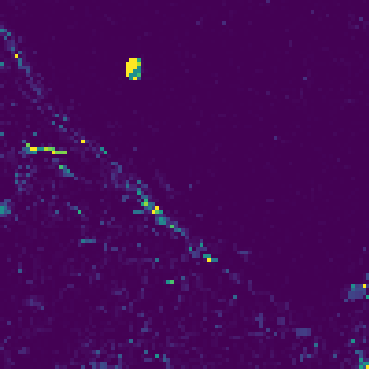}
\vspace{-0.3cm}

%\includegraphics[width=13mm,height=13mm]{picture/EAS/EAS_RX_AB4.png}
%\vspace{-0.3cm}

%\includegraphics[width=13mm,height=13mm]{picture/AED/AA4.png}
%\vspace{-0.3cm}

%\includegraphics[width=13mm,height=10.4mm]{picture/EAS/EAS_HYDICE.png}
%\vspace{-0.3cm}
\end{minipage}
}%插入图片，[]中设置图片大小，{}中是图片文件名\
\subfigure[]{
\label{CNNGAN}
\begin{minipage}[t]{0.065\linewidth}
\centering
%\includegraphics[width=13mm,height=13mm]{picture/AEGAN/AU2.png}
%\vspace{-0.3cm}

%\includegraphics[width=13mm,height=13mm]{picture/CNNGAN/AU3.png}
%\vspace{-0.3cm}

\includegraphics[width=13mm,height=13mm]{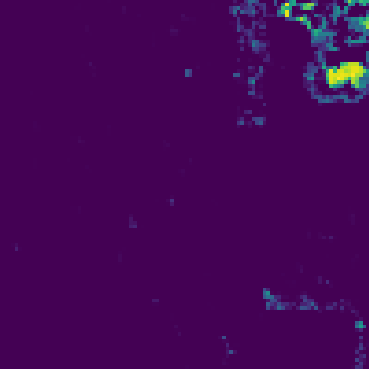}
\vspace{-0.3cm}

\includegraphics[width=13mm,height=13mm]{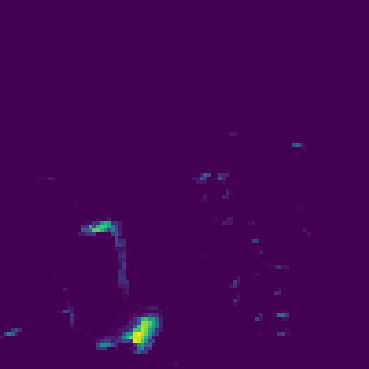}
\vspace{-0.3cm}

\includegraphics[width=13mm,height=13mm]{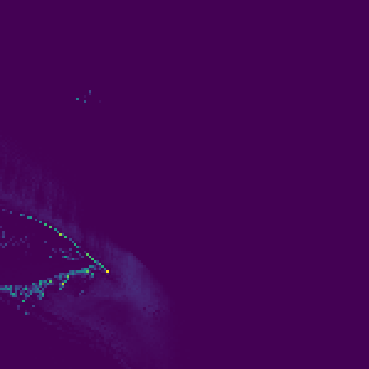}
\vspace{-0.3cm}

\includegraphics[width=13mm,height=13mm]{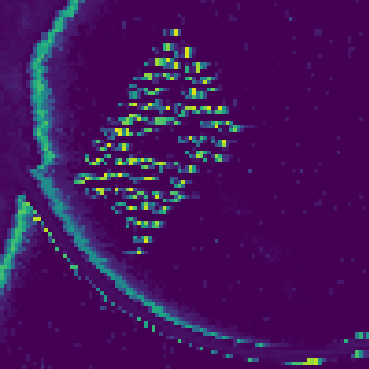}
\vspace{-0.3cm}

\includegraphics[width=13mm,height=13mm]{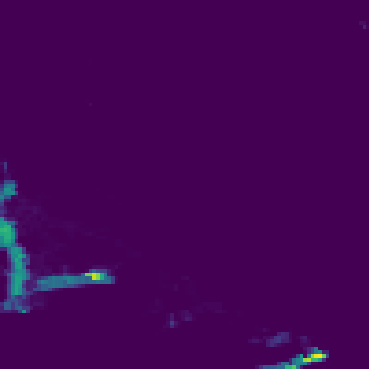}
\vspace{-0.3cm}

%\includegraphics[width=13mm,height=13mm]{picture/AEGAN/AB4.png}
%\vspace{-0.3cm}

%\includegraphics[width=13mm,height=13mm]{picture/CNNGAN/AA4.png}
%\vspace{-0.3cm}

%\includegraphics[width=13mm,height=10.4mm]{picture/AEGAN/AEGAN_HYDICE.png}
%\vspace{-0.3cm}
\end{minipage}
}%插入图片，[]中设置图片大小，{}中是图片文件名\
\subfigure[]{
\label{AEGAN}
\begin{minipage}[t]{0.065\linewidth}
\centering
%\includegraphics[width=13mm,height=13mm]{picture/AUTO/AU2.png}
%\vspace{-0.3cm}

%\includegraphics[width=13mm,height=13mm]{picture/AEGAN/AU3.png}
%\vspace{-0.3cm}

\includegraphics[width=13mm,height=13mm]{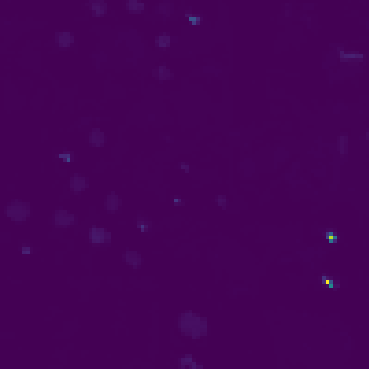}
\vspace{-0.3cm}

\includegraphics[width=13mm,height=13mm]{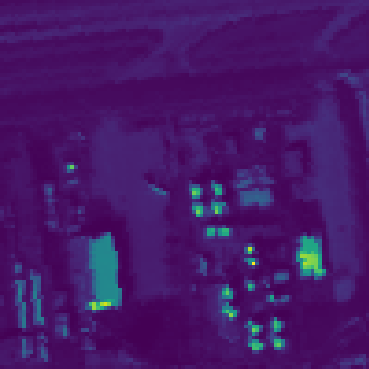}
\vspace{-0.3cm}

\includegraphics[width=13mm,height=13mm]{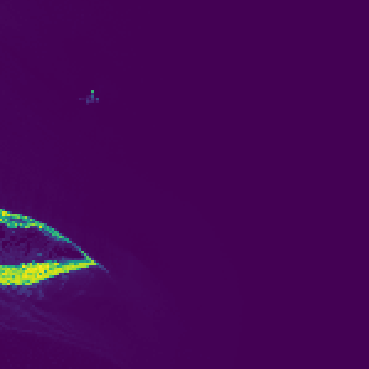}
\vspace{-0.3cm}

\includegraphics[width=13mm,height=13mm]{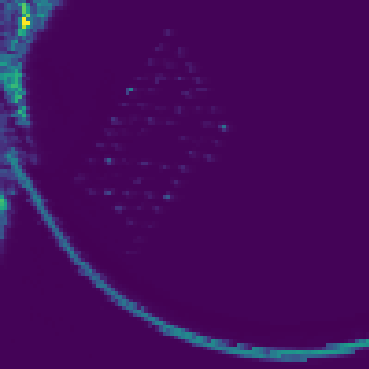}
\vspace{-0.3cm}

\includegraphics[width=13mm,height=13mm]{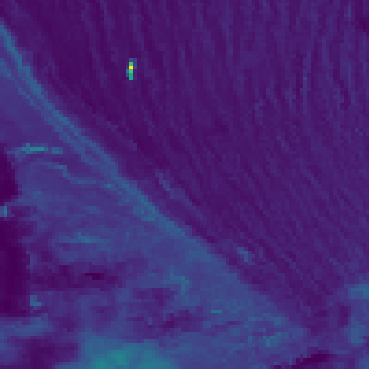}
\vspace{-0.3cm}

%\includegraphics[width=13mm,height=13mm]{picture/AUTO/AB4.png}
%\vspace{-0.3cm}

%\includegraphics[width=13mm,height=13mm]{picture/AEGAN/AA4.png}
%\vspace{-0.3cm}

%\includegraphics[width=13mm,height=10.4mm]{picture/AUTO/AUTO_HYDICE.png}
%\vspace{-0.3cm}
\end{minipage}
}%插入图片，[]中设置图片大小，{}中是图片文件名\
\subfigure[]{
\label{RGAE}
\begin{minipage}[t]{0.065\linewidth}
\centering
%\includegraphics[width=13mm,height=13mm]{picture/RGAE/RGAE_AU2.png}
%\vspace{-0.3cm}

%\includegraphics[width=13mm,height=13mm]{picture/PAB-DC/AU3.png}
%\vspace{-0.3cm}

\includegraphics[width=13mm,height=13mm]{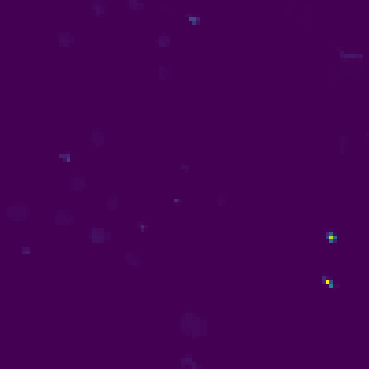}
\vspace{-0.3cm}

\includegraphics[width=13mm,height=13mm]{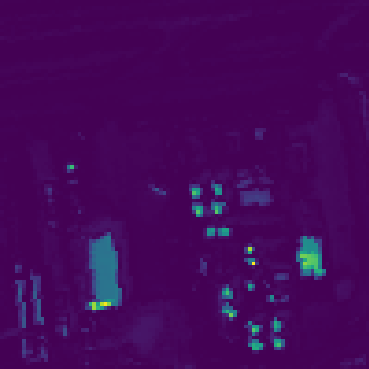}
\vspace{-0.3cm}

\includegraphics[width=13mm,height=13mm]{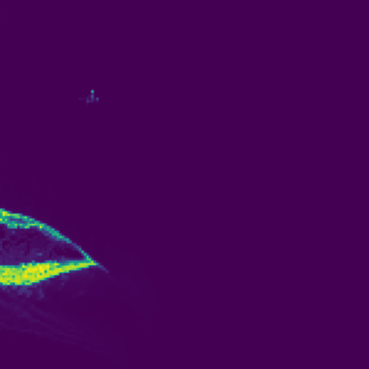}
\vspace{-0.3cm}

\includegraphics[width=13mm,height=13mm]{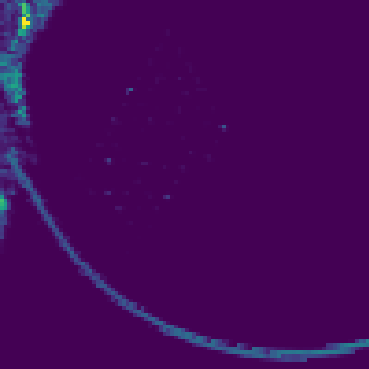}
\vspace{-0.3cm}

\includegraphics[width=13mm,height=13mm]{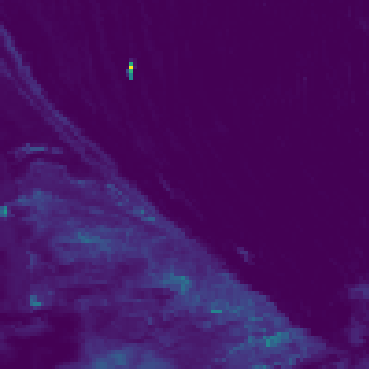}
\vspace{-0.3cm}

%\includegraphics[width=13mm,height=13mm]{picture/RGAE/RGAE_AB4.png}
%\vspace{-0.3cm}

%\includegraphics[width=13mm,height=13mm]{picture/PAB-DC/AA4.png}
%\vspace{-0.3cm}

%\includegraphics[width=13mm,height=10.4mm]{picture/RGAE/RGAE_HYDICE.png}
%\vspace{-0.3cm}
\end{minipage}
}%插入图片，[]中设置图片大小，{}中是图片文件名\
\subfigure[]{
\label{GAED}
\begin{minipage}[t]{0.065\linewidth}
\centering
%\includegraphics[width=13mm,height=13mm]{picture/GAED/GAED_AU2.png}
%\vspace{-0.3cm}

%\includegraphics[width=13mm,height=13mm]{picture/AUTO/AU3.png}
%\vspace{-0.3cm}

\includegraphics[width=13mm,height=13mm]{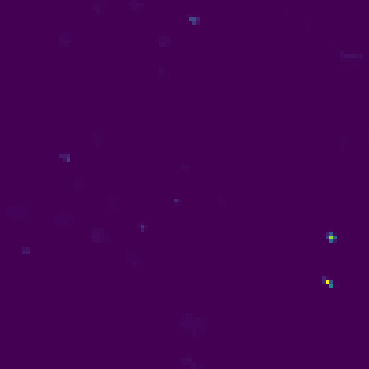}
\vspace{-0.3cm}

\includegraphics[width=13mm,height=13mm]{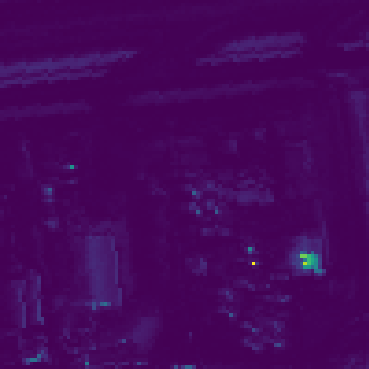}
\vspace{-0.3cm}

\includegraphics[width=13mm,height=13mm]{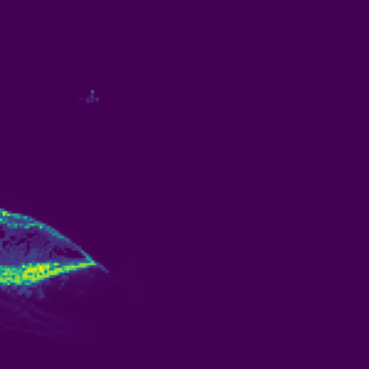}
\vspace{-0.3cm}

\includegraphics[width=13mm,height=13mm]{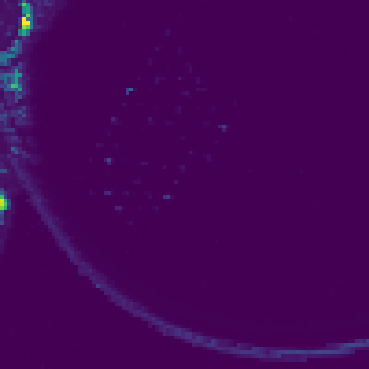}
\vspace{-0.3cm}

\includegraphics[width=13mm,height=13mm]{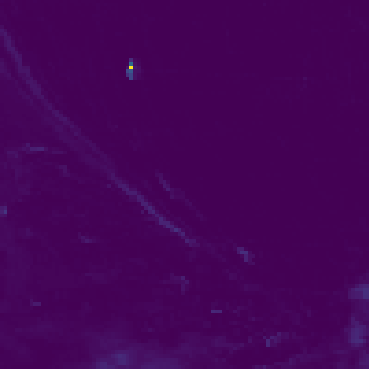}
\vspace{-0.3cm}

%\includegraphics[width=13mm,height=13mm]{picture/GAED/GAED_AB4.png}
%\vspace{-0.3cm}

%\includegraphics[width=13mm,height=13mm]{picture/AUTO/AA4.png}
%\vspace{-0.3cm}

%\includegraphics[width=13mm,height=10.4mm]{picture/GAED/GAED_HYDICE.png}
%\vspace{-0.3cm}
\end{minipage}
}%插入图片，[]中设置图片大小，{}中是图片文件名\
\subfigure[]{
\label{CaGAN}
\begin{minipage}[t]{0.065\linewidth}
\centering
%\includegraphics[width=13mm,height=13mm]{picture/CaGAN/AU2.png}
%\vspace{-0.3cm}

%\includegraphics[width=13mm,height=13mm]{picture/CaGAN/AU3.png}
%\vspace{-0.3cm}

\includegraphics[width=13mm,height=13mm]{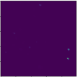}
\vspace{-0.3cm}

\includegraphics[width=13mm,height=13mm]{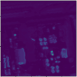}
\vspace{-0.3cm}

\includegraphics[width=13mm,height=13mm]{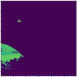}
\vspace{-0.3cm}

\includegraphics[width=13mm,height=13mm]{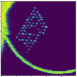}
\vspace{-0.3cm}

\includegraphics[width=13mm,height=13mm]{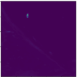}
\vspace{-0.3cm}

%\includegraphics[width=13mm,height=13mm]{picture/CaGAN/AB4.png}
%\vspace{-0.3cm}

%\includegraphics[width=13mm,height=13mm]{picture/CaGAN/AA4.png}
%\vspace{-0.3cm}

%\includegraphics[width=13mm,height=10.4mm]{picture/CaGAN/CAGAN_HYDICE.png}
%\vspace{-0.3cm}
\end{minipage}
}%插入图片，[]中设置图片大小，{}中是图片文件名\

\caption{Visualization of the detection results, the data set from up to down is Los Angeles-1, Los Angeles-2, Cat Island, San Diego, Bay Champagne. (a) Color composites of HSI. (b) Groundtruth map. (c) RX. (d) LRX. (e) PAB. (f) AED. (g) EAS-RX. (h) AEGAN. (i) Auto-AD. (j) RGAE. (k) GAED. (l) CaGAN.} %最终文档中希望显示的图片标题htb
\label{detectionCaGAN} %用于文内引用的标签
\end{figure*}

\begin{figure*}[!ht] %H为当前位置，!htb为忽略美学标准，htbp为浮动图形
\centering %图片居中

\subfigure[]{
\label{ROCAU2}
\begin{minipage}[t]{0.24\linewidth}
\centering
\hspace{-6mm}
\includegraphics[width=47mm,height=37mm]{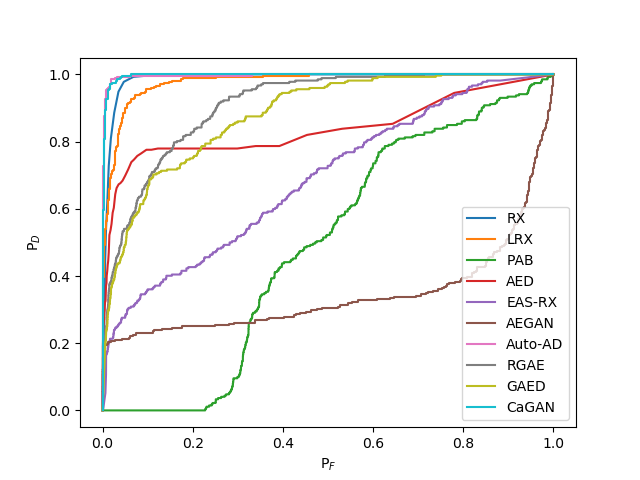}
\end{minipage}
}%插入图片，[]中设置图片大小，{}中是图片文件名
\subfigure[]{
\label{ROCAU3}
\begin{minipage}[t]{0.24\linewidth}
\centering
\hspace{-6mm}
\includegraphics[width=47mm,height=37mm]{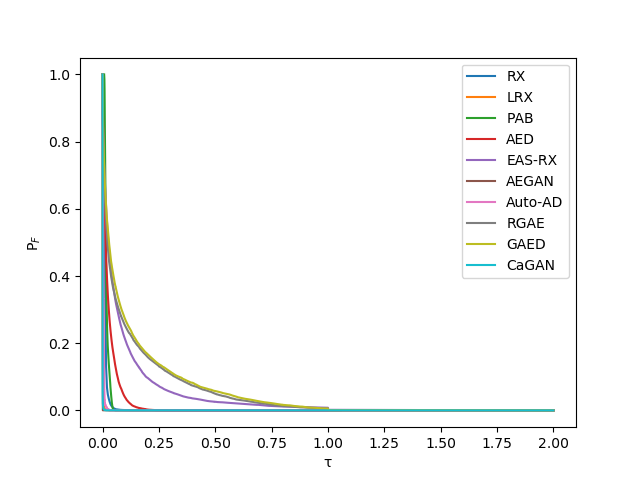}
\end{minipage}
}%插入图片，[]中设置图片大小，{}中是图片文件名
\subfigure[]{
\label{ROCAU4}
\begin{minipage}[t]{0.24\linewidth}
\centering 
\hspace{-6mm}
\includegraphics[width=47mm,height=37mm]{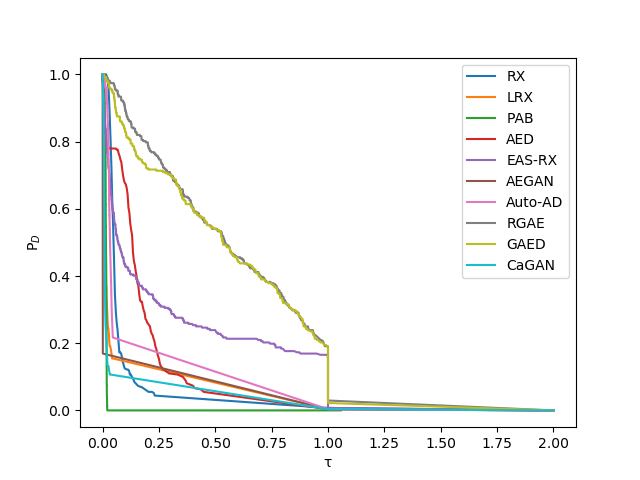}
\end{minipage}
}%插入图片，[]中设置图片大小，{}中是图片文件名
\subfigure[]{
\label{ROCAU5}
\begin{minipage}[t]{0.24\linewidth}
\centering
\hspace{-6mm}
\includegraphics[width=47mm,height=37mm]{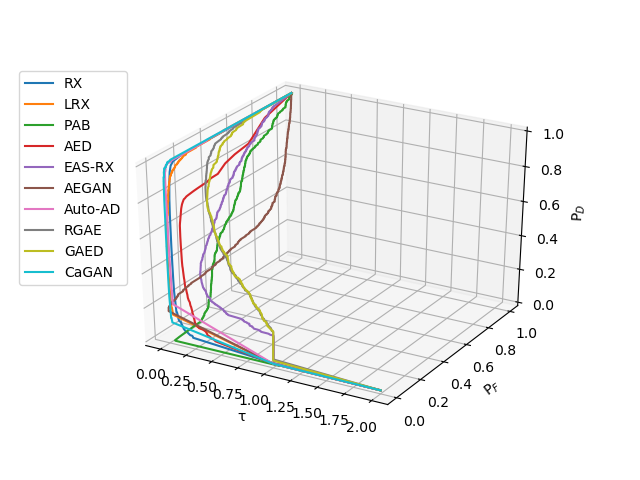}
\end{minipage}
}%插入图片，[]中设置图片大小，{}中是图片文件名

\subfigure[]{
\label{ROCAB1}
\begin{minipage}[t]{0.24\linewidth}
\centering
\hspace{-6mm}
\includegraphics[width=47mm,height=37mm]{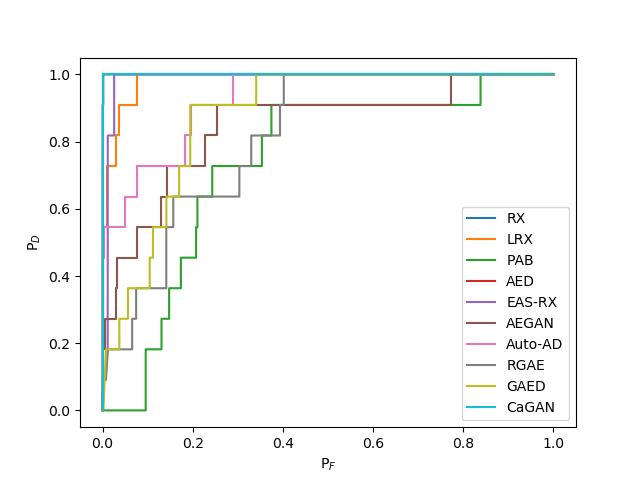}
\end{minipage}
}%插入图片，[]中设置图片大小，{}中是图片文件名
\subfigure[]{
\label{ROCAB2}
\begin{minipage}[t]{0.24\linewidth}
\centering
\hspace{-6mm}
\includegraphics[width=47mm,height=37mm]{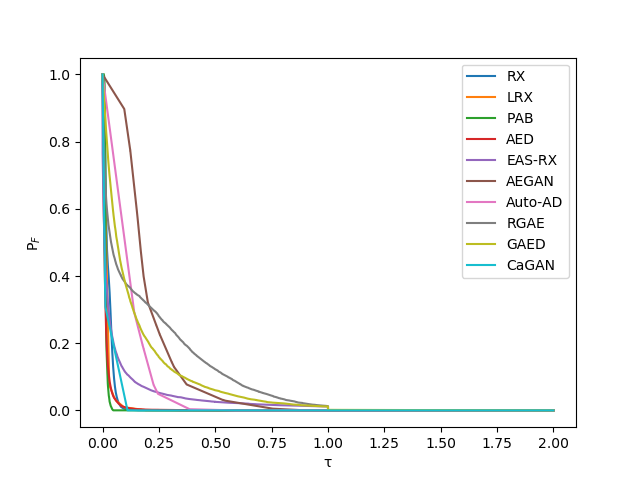}
\end{minipage}
}%插入图片，[]中设置图片大小，{}中是图片文件名
\subfigure[]{
\label{ROCAB3}
\begin{minipage}[t]{0.24\linewidth}
\centering
\hspace{-6mm}
\includegraphics[width=47mm,height=37mm]{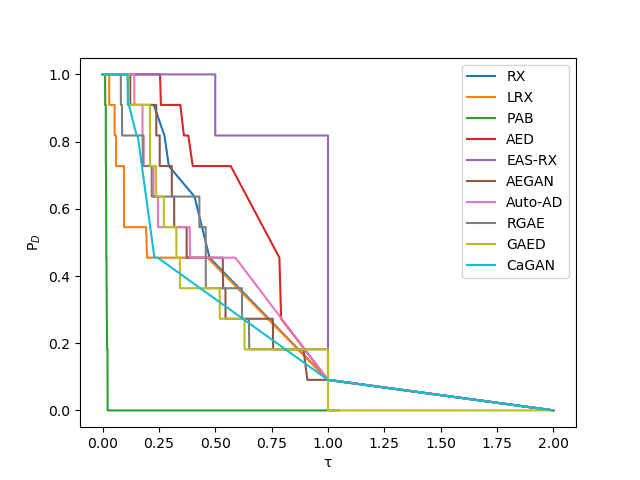}
\end{minipage}
}%插入图片，[]中设置图片大小，{}中是图片文件名
\subfigure[]{
\label{ROCAB4}
\begin{minipage}[t]{0.24\linewidth}
\centering
\hspace{-6mm}
\includegraphics[width=47mm,height=37mm]{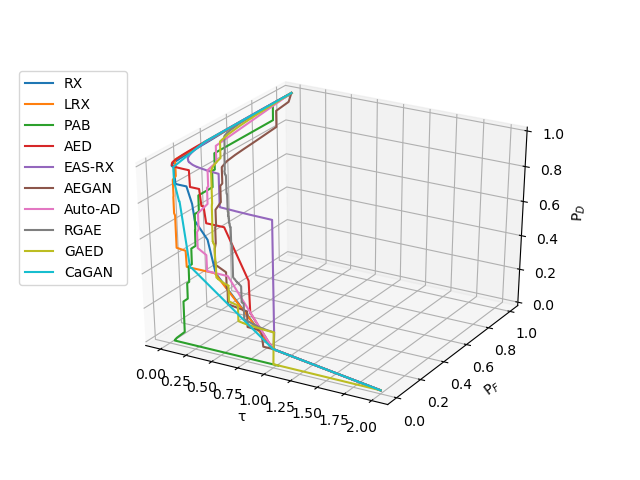}
\end{minipage}
}%插入图片，[]中设置图片大小，{}中是图片文件名

\caption{Eight ROC curves of Los Angeles-1 and Bay Champagne datasets using different methods: (a) 2D ROC Curves ($P_D$, $P_F$) of Los Angeles-1. (b) 2D ROC Curves ($P_F$, $\tau$) of Los Angeles-1. (c) 2D ROC Curves ($P_D$, $\tau$) of Los Angeles-1. (d) 3D ROC Curves of Los Angeles-1. (e) 2D ROC Curves ($P_D$, $P_F$) of Bay Champagne. (f) 2D ROC Curves ($P_F$, $\tau$) of Bay Champagne. (g) 2D ROC Curves ($P_D$, $\tau$) of Bay Champagne. (h) 3D ROC Curves of Bay Champagne.} %最终文档中希望显示的图片标题htb
\label{ROC} %用于文内引用的标签
\end{figure*}

\begin{table*}[ht]
	\centering
	\caption{Evaluation $\textbf{AUC}_{(D,F)}$ Scores Obtained From Different Methods For Different Datasets}
	\label{AUCtable}
	\setlength{\tabcolsep}{1.5mm}{
	\begin{tabular}{ccccccccccc}
		\toprule  % 顶部线
		HSIs &	RX & LRX & PAB &	AED & EAS-RX &	 AEGAN & Auto-AD & 	RGAE &	GAED & CaGAN \\
		%\midrule  % 中部线
		%Texas Coast &0.9942	&0.9770&	0.9929&	0.8145&	0.9108	&0.5363&	0.9988	&0.9993	&0.9970&	\textbf{0.9989} \\
		\midrule  % 中部线
		Los Angeles-1 &	0.9884&	0.9777&	0.4920&	0.8385&	0.8021&	0.3432&	0.9965	&0.9948&	0.9931&	\textbf{0.9965}\\
		\midrule  % 中部线
		Los Angeles-2 &	0.9693&	0.8526&	0.4878	&0.9642&	0.8259&	0.4770	&0.9620&	0.9597&	0.9044	&\textbf{0.9741}\\
		\midrule  % 中部线
		Cat Island &0.9807&	0.9808&	0.6495&	0.9846&	0.9500&	0.8527&	0.9510	&0.9501	&0.9163	&\textbf{0.9860}\\
		\midrule  % 中部线
		San Diego &	0.9104&	0.8884&	0.8941&	\textbf{0.9413}&	0.8346&	0.8458&	0.9042	&0.9020&	0.8030&	0.8930 \\
		\midrule  % 中部线
		Bay Champagne &	0.9999&	0.9855&	0.7390&	0.9997&	0.9986&	0.8483&	0.9276	&0.8651&	0.9872&	\textbf{0.9999} \\
		%\midrule  % 中部线
		%Pavia &	0.9535&	0.9460&	0.6334&	0.9840&	0.9367&	0.7712&	0.9867&	0.9075	&0.9380	& \textbf{0.9936} \\
		%\midrule  % 中部线
		%HYDICE & 0.9855&	0.8217&	0.8371&	0.9799&	0.9733&	0.3164&	0.9821&	0.8307&	0.9620&\textbf{0.9960} \\
		\midrule  % 中部线https://www.overleaf.com/project/62d66ee3cf6e214a68ed1a93
		Average &	0.9643&	0.9256	&0.6478&	0.9457&	0.8633&	0.6285	&0.9555&	0.9466&	0.9109	&\textbf{0.9658}\\
        \midrule  % 中部线
		{test(min) } &	{ 5.327}	& {9.182} &	{ 0.125}	& {0.035} &	{0.234} &	{0.522} &	{0.95} &	{6.526} &	{2.815}	& {0.795}\\
        \midrule  % 中部线
		{Parameters} &	{-}	& {-} &	{-}	& {-} &	{-} & {3534852} &{3189538} &	{35275}& {54030}& {1140426}\\
		\bottomrule  % 底部线
	\end{tabular}
	}
	
\end{table*}

\begin{table*}[!ht]
	\centering
	\caption{Detailed AUC Results with Different Methods for Los Angeles-1 Datasets}
	\label{result1}
	\setlength{\tabcolsep}{1.5mm}{
	\begin{tabular}{cccccccccc}           
		\toprule  % 顶部线
         &method & $\textbf{AUC}_{(D,F)} $& $\textbf{AUC}_{(D,\tau)}$&	$\textbf{AUC}_{(F,\tau)}$	& $\textbf{AUC}_{TD}$& $\textbf{AUC}_{BS}$	& $\textbf{AUC}_{TDBS}$& $\textbf{AUC}_{SNPR}$ & $\textbf{AUC}_{ODP}$ \\
		\midrule  % 中部线
		\multirow{5}{*}{Traditional-based}&RX &0.9884&	0.0893&	0.0115&	\textbf{1.0777}&	0.9769&	0.0778&	7.7652&	\textbf{1.0662}\\
  	  %\midrule  % 中部线
		&LRX &0.9777&0.0295&	0.0006&	1.0072&	0.9771&	0.0289&	49.1667&1.0066\\
		%\midrule  % 中部线
           & PAB&0.4920&	0.1580&	0.2077&	0.6500&	0.2843&	-0.0497&0.7607&	0.4423\\
            %\midrule  % 中部线
            &AED&0.8385&\textbf{0.1622}&0.0256&1.0007&0.8129&	0.1366&	6.3359&	0.9751\\
		%\midrule  % 中部线
            &EAS-RX&0.8021&0.1593&	0.0100&	0.9614&	0.7921&\textbf{0.1493}&	15.9300&	0.9514\\ 
		\midrule  % 中部线
		\multirow{5}{*}{DL-based}&AEGAN &	0.3432&	\textbf{0.0082}&	0.0011&	0.3514&	0.3421&	0.0071&	7.4545&	0.3503\\
		%\midrule  % 中部线
		&Auto-AD &0.9965&	0.0555&	0.0038&	1.0520&	0.9927&	0.0517&	14.6052&	1.0482\\
		%\midrule  % 中部线
		&RGAE &	0.9948&	0.0389&	0.0027&	1.0337&	0.9921&	0.0362&	14.4074&1.0310\\
		%\midrule  % 中部线
		&GAED &0.9931&	0.0321&	0.0004&	1.0252&	0.9927&	0.0317&	80.2500&	1.0248\\
           %\midrule  % 中部线
		&CaGAN &\textbf{0.9965}&	0.0256&	\textbf{0.0002}&	1.0221&\textbf{0.9963}&	0.0254&	\textbf{160}&	1.0219\\
		\bottomrule  % 底部线
	\end{tabular}
	}
	
\end{table*}
\begin{table*}[!ht]
	\centering
	\caption{Detailed AUC Results with Different Methods for Bay Champagne Datasets}
	\label{result2}
	\setlength{\tabcolsep}{1.5mm}{
	\begin{tabular}{cccccccccc}           
		\toprule  % 顶部线
           &method & $\textbf{AUC}_{(D,F)} $& $\textbf{AUC}_{(D,\tau)}$&	$\textbf{AUC}_{(F,\tau)}$	& $\textbf{AUC}_{TD}$& $\textbf{AUC}_{BS}$	& $\textbf{AUC}_{TDBS}$& $\textbf{AUC}_{SNPR}$ & $\textbf{AUC}_{ODP}$  \\
		\midrule  % 中部线
		\multirow{5}{*}{Traditional-based}&RX &0.9999&	0.5314&	0.0260&	1.5313&	0.9739&	0.5054&	20.4385&	1.5053\\
  	  %\midrule  % 中部线
		&LRX &0.9855&	0.3633&	0.0083&	1.3488&	0.9772&	0.3550&	43.7711&1.3405\\
		%\midrule  % 中部线
           & PAB&0.7390&	0.2808&	0.2283&	1.0198&	0.5107&	0.0525&	1.2300&	0.7915\\
            %\midrule  % 中部线
            &AED&0.9997&	0.6873&	0.0135&	1.6870&	0.9862&	0.6738&	50.9111&	1.6735\\
		%\midrule  % 中部线
            &EAS-RX&0.9986&	\textbf{0.9091}&	0.0174&	\textbf{1.9077}&	0.9812&	\textbf{0.8917}&	\textbf{52.2471}&	\textbf{1.8903}\\ 
		\midrule  % 中部线
		\multirow{5}{*}{DL-based}&AEGAN &0.8483&	0.4862&	0.2006&	1.3345&	0.6477&	0.2856&	2.4237&	1.1339\\
		%\midrule  % 中部线
		&Auto-AD &0.9276&	0.4692&	0.1141&	1.3968&	0.8135&	0.3551&	4.1122&	1.2827\\
		%\midrule  % 中部线
		&RGAE &	0.8651&		0.3599&		0.0449&		1.2250&		0.8202&		0.3150&		8.0156&		1.1801\\
		%\midrule  % 中部线
		&GAED &0.9872&	0.2766&	0.0099&	1.2638&	0.9773&	0.2667&	27.9394&	1.2539\\
           %\midrule  % 中部线
		&CaGAN &\textbf{0.9999}&	0.3400&	\textbf{0.0079}&	1.3399&	\textbf{0.9920}&	0.3321&	43.0380&	1.3320\\
		\bottomrule  % 底部线
	\end{tabular}
	}
	
\end{table*}

\subsection{Performance of CaGAN for HAD}

The detection results of specified HAD task, i.e., comparison methods, evaluation metrics and ablation study are presented and discussed in this section.

\subsubsection{Comparison Methods and Evaluation Metrics} 

We evaluate the effectiveness of the proposed algorithm with nine typical anomaly detection methods, i.e., the RX \cite{reed1990adaptive}, LRX \cite{2014A}, 
PAB\cite{huyan2018hyperspectral}, attribute and edge-preserving filters (AED)\cite{kang2017hyperspectral}, EAS-RX\cite{9740135},  AE-based GAN (AEGAN) , Auto-AD{\cite{9382262}}, GRAE{\cite{9494034}} and GAED{\cite{9893839}}. 

To quantitatively evaluate the performance of different detectors, the receiver operating characteristics (ROC) \cite{zweig1993receiver} and the area under the ROC curve (AUC) \cite{ferri2011coherent} are applied as performance indicators. Specifically, ROC describes the different relationships between the true positive rate ($P_D$) and the false positive rate ($P_F$). $P_D$ defines the proportion of correctly assigned positive results occurring in all positive samples, and $P_F$ defines the proportion of false positive results occurring in all results and vice versa for available negative samples. However, both $P_D$ and $P_F$ are calculated by the same thresholds used by the detector, therefore if only use $\textbf{AUC}_{(D,F)}$ to express 2D ROC is not credible \cite{9909988}. When $P_D$ and $P_F$ are very high, the calculated $\textbf{AUC}_{(D,F)}$ is also very high. Likewise, both $P_D$ and $P_F$ are very low, and related $\textbf{AUC}_{(D,F)}$ is also very low, that is, $P_D$ and $P_F$ are bundled together and cannot work independently.

An evaluation tool based on 3D ROC analysis that extends traditional 2D ROC analysis by including the threshold $\tau$ as an additional independent parameter to represent the 3D ROC curve as a function of the three parameters $P_D$, $P_F$ and $\tau$. The 3D ROC curve was developed to generate three 2D ROC curves to evaluate HAD in all aspects from eight detection measures \cite{chang2023exploration}. Therefore, a 3D ROC curve can be generated by a triplet parameter vector specified by ($P_D$, $P_F$, $\tau$), or by three 2D ROC curve of ($P_D$, $P_F$), ($P_D$, $\tau$) and ($P_F$, $\tau$) and their AUC values expressed as $\textbf{AUC}_{(D,F)}$, $\textbf{AUC}_{(D,\tau)}$ and $\textbf{AUC}_{(F,\tau)}$, where $\textbf{AUC}_{(D,\tau)}$ and $\textbf{AUC}_{(F,\tau)}$ can be used respectively for evaluating target detection TD and background suppression rate BS. In the experiment, we use $\textbf{AUC}_{(F,\tau)}$, $\textbf{AUC}_{BS}$ and $\textbf{AUC}_{SNPR}$ to represent the background suppression rate, and $\textbf{AUC}_{(D,F)}$ and $\textbf{AUC}_{ODP}$ are used to evaluate the performance of the detector. Apart from the three AUC values mentioned, the study conducted by \cite{chang2021orthogonal,9579465,9460785,9205919}, and \cite{9387095} introduced five novel AUC measures that have demonstrated high efficacy in quantifying various aspects of detection performance. Specifically, these measures assess joint target detection (TD), joint background suppression (BS), the combined metric of target detection and background suppression (TDBS), signal-to-noise probability ratio (SNPR), and overall detection performance (ODP). 
The specific calculation formula for each evaluation index is represented as follows.
\begin{equation}
\begin{split}
\label{eq188}
\textbf{AUC}_{TD} = {\textbf{AUC}_{(D,F)} + 
 \textbf{AUC}_{(D,\tau)}}
\end{split}
\end{equation}
\begin{equation}
\begin{split}
\label{eq189}
\textbf{AUC}_{BS} = {\textbf{AUC}_{(D,F)} - 
 \textbf{AUC}_{(F,\tau)}}
\end{split}
\end{equation}
\begin{equation}
\begin{split}
\label{eq190}
\textbf{AUC}_{TDBS} = {\textbf{AUC}_{(D,\tau)} - 
 \textbf{AUC}_{(F,\tau)}}
\end{split}
\end{equation}
\begin{equation}
\begin{split}
\label{eq191}
\textbf{AUC}_{SNPR} = {\textbf{AUC}_{(D,\tau)} / 
 \textbf{AUC}_{(F,\tau)}}
\end{split}
\end{equation}
\begin{equation}
\begin{split}
\label{eq192}
\textbf{AUC}_{ODP} = {\textbf{AUC}_{(D,F)} + 
 \textbf{AUC}_{(D,\tau)} -\textbf{AUC}_{(F,\tau)} }
\end{split}
\end{equation}

The detection maps obtained from different anomaly detection comparison methods on the five experimental datasets are presented in Fig.\ref{detectionCaGAN}. The detection maps reveal that our algorithm can extract more distinguishable features and maximally detect anomalies with the influence in the noisy areas. Compared with other comparison methods, the CaGAN achieves the most comparable visual results on all other datasets especially for the Cat Island and Bay Champagne in the third and fifth row in Fig.\ref{detectionCaGAN}. Meanwhile, among all comparison methods, we can observe that PAB and AEGAN can detect anomalies clearly but suffer from false alarms. On the contrary, the LRX, Auto-AD, RGAE and GAED yield fewer false alarms. However, they neglect some anomalies resulting in a low rate of detection and irregular target shapes. RX, AED and EAS-RX detect a few anomalies with low confidence and provide more distinguishable detection results, while some anomalous target detected by the RX and LRX cannot be preserved completely. Due to lack of spatial location information, the detection maps obtained by AEGAN presented a lot of random noises for anomaly detection. The proposed CaGAN reaches an exquisite balance between the high detection rate and the low false alarm rate. For Cat Island and San Diego data, CaGAN structure can detect most anomalies as well as preserving integral target shapes. Meanwhile, observing from Los Angeles-1, Los Angeles-2 and Bay Champagne data, the proposed CaGAN also provides the most distinguishable detection results with the fewest false alarms compared with other comparison methods.

ROC curves for HSIs of Los Angeles-1 and Bay Champagne are illustrated in Fig.\ref{ROC}. Three 2D ROC curves were generated for performance evaluation, specifically the 2D ROC curve of ($P_F$, $\tau$) to be used to evaluate the degree and level of BS that an anomaly detector can achieve. The curve charts in Fig.\ref{ROCAU2} and Fig.\ref{ROCAB1} represent the value of 2D ROC calculated using $P_D$ and $P_F$. The larger the value, the better for the method to detect anomalies. The ROC value of our proposed CaGAN indicates the largest value than other methods. Fig.\ref{ROCAU3} and Fig.\ref{ROCAB2} represents the value of 2D ROC calculated using $P_F$ and $\tau$, CaGAN presents the smallest value, that is, the background suppression effect is the best for the proposed CaGAN. Fig.\ref{ROCAU4} and Fig.\ref{ROCAB3} represent the value of 2D ROC calculated using $P_D$ and $\tau$, which reflect the anomaly detection effect. In addition, in Fig.\ref{ROCAU5} and Fig.\ref{ROCAB4}, 3D ROC curves for Los Angeles-1 and Bay Champagne also indicate the performance for the proposed CaGAN in a more comprehensive way. In general, the proposed CaGAN presents the best performance for BS and detection among all comparison methods in terms of the ROC curves.

The $\textbf{AUC}_{(D,F)}$ score is presented in Table \ref{AUCtable}, it intuitively compares the performance of different detectors (The optimal scores are in bold). Especially for Cat Island and Bay Champagne datasets, we can observe that these datasets only contain a small number of anomalies and all the methods present good detection performance, while the proposed CaGAN demonstrates the best $\textbf{AUC}_{(D,F)}$ scores. For other HSI datasets, the proposed CaGAN exhibits more robust and satisfying results than other comparable methods. Meanwhile, it presents the best detection accuracy for most of datasets with the highest average of $\textbf{AUC}_{(D,F)}$.

The specific AUC values for Los Angeles-1 and Bay Champagne are illustrated in Table \ref{result1} and Table \ref{result2}. The different numerical results of nine comparison methods under eight evaluation indicators ($\textbf{AUC}_{(D,F)}$, $\textbf{AUC}_{(D,\tau)}$, $\textbf{AUC}_{(F,\tau)}$, $\textbf{AUC}_{TD}$, $\textbf{AUC}_{BS}$, $\textbf{AUC}_{TDBS}$, $\textbf{AUC}_{SNPR}$ and $\textbf{AUC}_{ODP}$) are shown in these Tables. According to the comparison results of $\textbf{AUC}_{(F,\tau)}$, $\textbf{AUC}_{BS}$ and $\textbf{AUC}_{SNPR}$, we can observe that the proposed CaGAN also indicates the optimal effect in background suppression.

\subsubsection{Ablation Analysis of CaGAN}

In this part, we mainly analysis the effect of CBM and the reconstruction results. 

(i) Effect Analysis of CBM

The CBM is essential to construct a pure training set for DL-based background estimation, which refrains the model from being contaminated by anomalies. The ablation analysis of CBM is presented in Table \ref{CBM}, we can conclude that the detection performance is greatly improved by CBM.

\begin{table}[!ht]
	\centering
	\caption{Ablation analysis of CBM. D-CBM stands for not using CBM}
	\label{CBM}
	\setlength{\tabcolsep}{3.5mm}{
	\begin{tabular}{ccc}
		\toprule  % 顶部线
		HSIs &	CBM	& D-CBM \\
		%\midrule  % 中部线
		%Texas Coast &	\textbf{0.9993} &	0.9951 \\
		\midrule  % 中部线
		Los Angeles-1 &	\textbf{0.9967} &	0.9790 \\
		\midrule  % 中部线
		Los Angeles-2 &	\textbf{0.9754} &	0.9522 \\
		\midrule  % 中部线
		Cat Island &	\textbf{0.9912} &	0.9244\\
		\midrule  % 中部线
		San Diego &	\textbf{0.9015} &	0.8703 \\
		\midrule  % 中部线
		Bay Champagne &	\textbf{0.9999} &	0.9996 \\
		%\midrule  % 中部线
		%Pavia &	\textbf{0.9988} &	0.9841 \\
		%\midrule  % 中部线
		%HYDICE &	\textbf{0.9960} &	0.9912 \\
		\bottomrule  % 底部线
	\end{tabular}
	}
	
\end{table}

\begin{figure}[!ht] %H为当前位置，!htb为忽略美学标准，htbp为浮动图形
\centering %图片居中

\subfigure[]{
\label{BoxAU2}
\begin{minipage}[t]{0.5\linewidth}
\centering
\includegraphics[width=45mm,height=40mm]{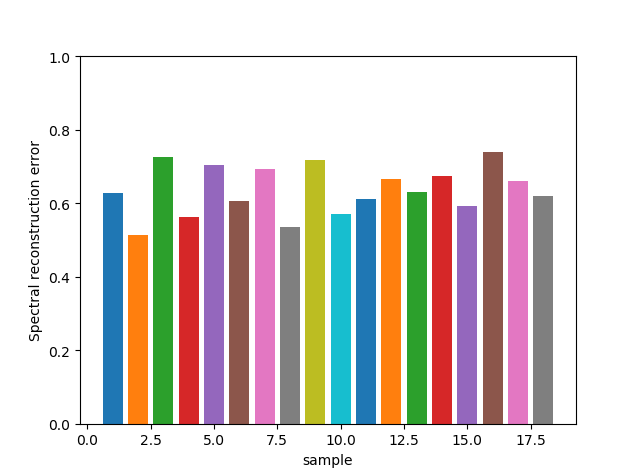}
\end{minipage}
}%插入图片，[]中设置图片大小，{}中是图片文件名
\subfigure[]{
\label{BoxAU4}
\begin{minipage}[t]{0.5\linewidth}
\centering
\includegraphics[width=45mm,height=40mm]{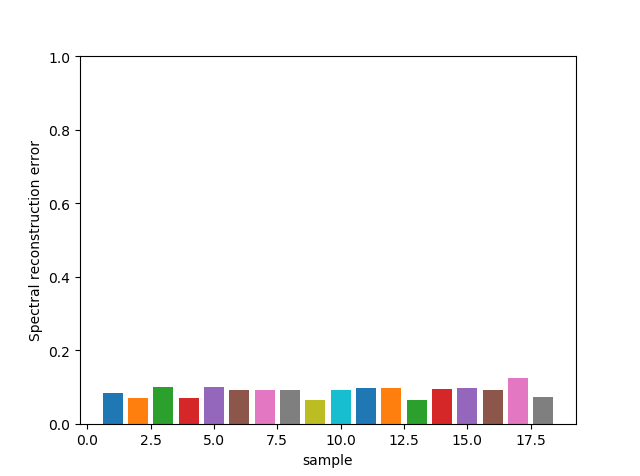}
\end{minipage}
}%插入图片，[]中设置图片大小，{}中是图片文件名

\caption{(a) Spectral reconstruction error of anomaly samples in Cat Island data. (b) Spectral reconstruction error of background samples in Cat Island data.} %最终文档中希望显示的图片标题htb
\label{RecBA} %用于文内引用的标签
\end{figure}

\begin{figure}[htb] %H为当前位置，!htb为忽略美学标准，htbp为浮动图形
\centering %图片居中
\includegraphics[scale=0.5]{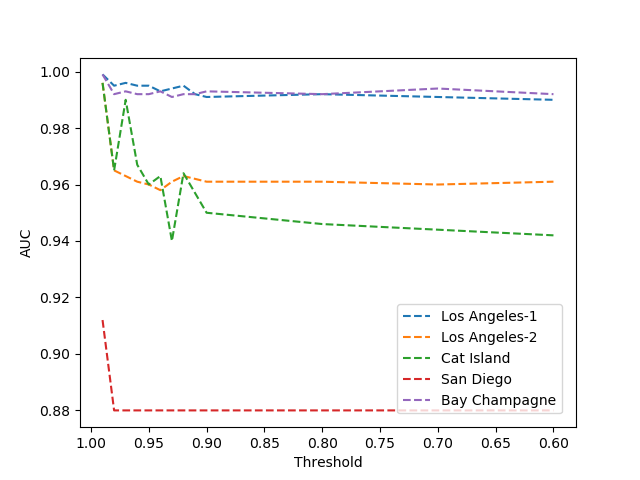} %插入图片，[]中设置图片大小，{}中是图片文件名
\caption{Effect Analysis of the threshold $\beta$ for CBM.} %最终文档中希望显示的图片标题htb
\label{threshold} %用于文内引用的标签
\end{figure}

(ii) Effect Analysis of Reconstruction

\begin{table*}[ht]
	\centering
	\caption{Continual Hyperspectral Anomaly Detection performance with two evaluation metrics.}
	\label{ACCBWT}
	\setlength{\tabcolsep}{3.5mm}{
	\begin{tabular}{ccccccccc}
		\toprule  % 顶部线
		\multirow{2}*{Method} &	\multicolumn{2}{c}{1-2 Tasks}	& \multicolumn{2}{c}{1-3 Tasks} & \multicolumn{2}{c}{1-4 Tasks} & \multicolumn{2}{c}{1-5 Tasks} \\
		\midrule  % 中部线
		~ & ACC &	BWT &	ACC	& BWT &	ACC &	BWT &	ACC	& BWT\\
		\midrule  % 中部线
		CaGAN &	0.5087	&	- & 0.7222 &	-	& 0.7881 &	- &	0.6246&-	 \\
		\midrule  % 中部线
		J-CaGAN &	0.9746	&	- & 0.9759 &	-	& 0.9100 &	- &	0.9115&-	 \\
		\midrule  % 中部线
		FT-CaGAN &	0.8478 &	-0.2612 &	0.9482 &	-0.0556 &	0.7537 &	-0.2228 &	0.7538 &	-0.2169	 \\
		\midrule  % 中部线
		D-CaGAN	& 0.8431 &	-0.2465 &	0.8210	& -0.2452 &	0.8387 &	-0.1528	& 0.7873 &	-0.5231 \\
		\midrule  % 中部线
		R-CaGAN &	0.9546 &	-0.0461	& 0.8942 &	-0.1363	& 0.7291 &	-0.1974 &	0.9439 &	-0.0244\\
		\midrule  % 中部线
		CL-CaGAN &	\textbf{0.9766}	& \textbf{0.0013} &	\textbf{0.9797} &	\textbf{-0.0058}	& \textbf{0.9153} &	\textbf{-0.0519} &	\textbf{0.9577} &	\textbf{-0.0031}\\
		
		\bottomrule  % 底部线
	\end{tabular}
	}
	
\end{table*}

\begin{figure*}[ht] %H为当前位置，!htb为忽略美学标准，htbp为浮动图形
\centering %图片居中
\subfigure[]{
\label{AUC}
\begin{minipage}[t]{0.29\linewidth}
\centering
\includegraphics[width=48mm,height=40mm]{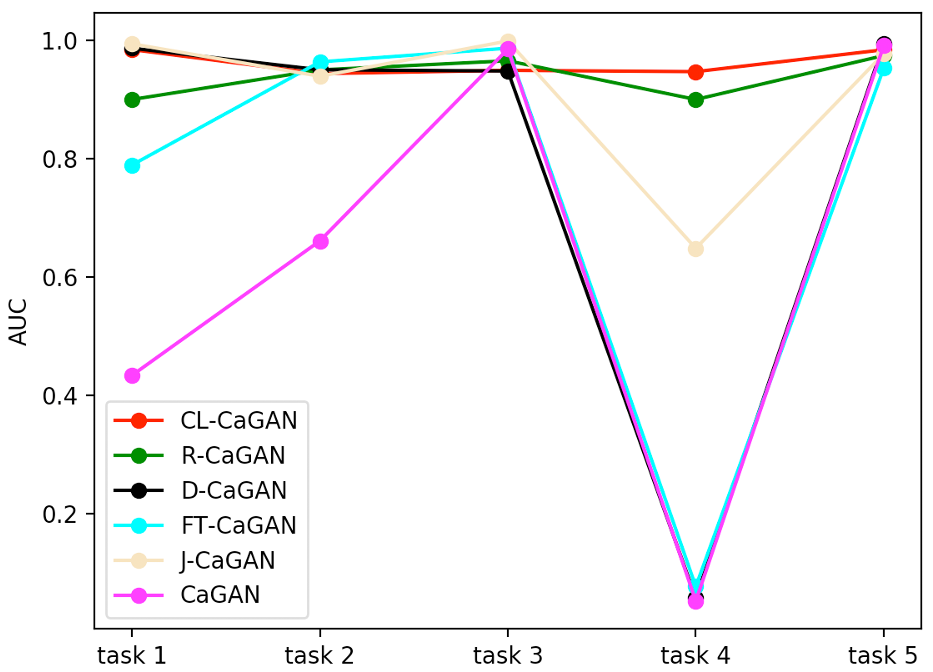}
\end{minipage}
}%插入图片，[]中设置图片大小，{}中是图片文件名
\subfigure[]{
\label{ACC}
\begin{minipage}[t]{0.29\linewidth}
\centering
\includegraphics[width=48mm,height=40mm]{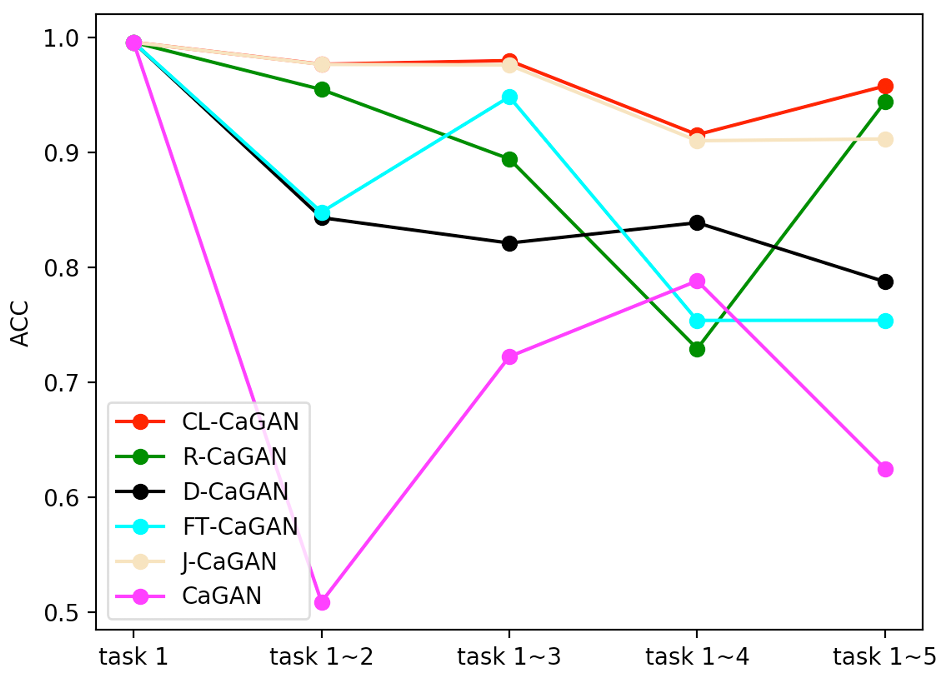}
\end{minipage}
}%插入图片，[]中设置图片大小，{}中是图片文件名
\subfigure[]{
\label{BWT}
\begin{minipage}[t]{0.29\linewidth}
\centering
\includegraphics[width=48mm,height=40mm]{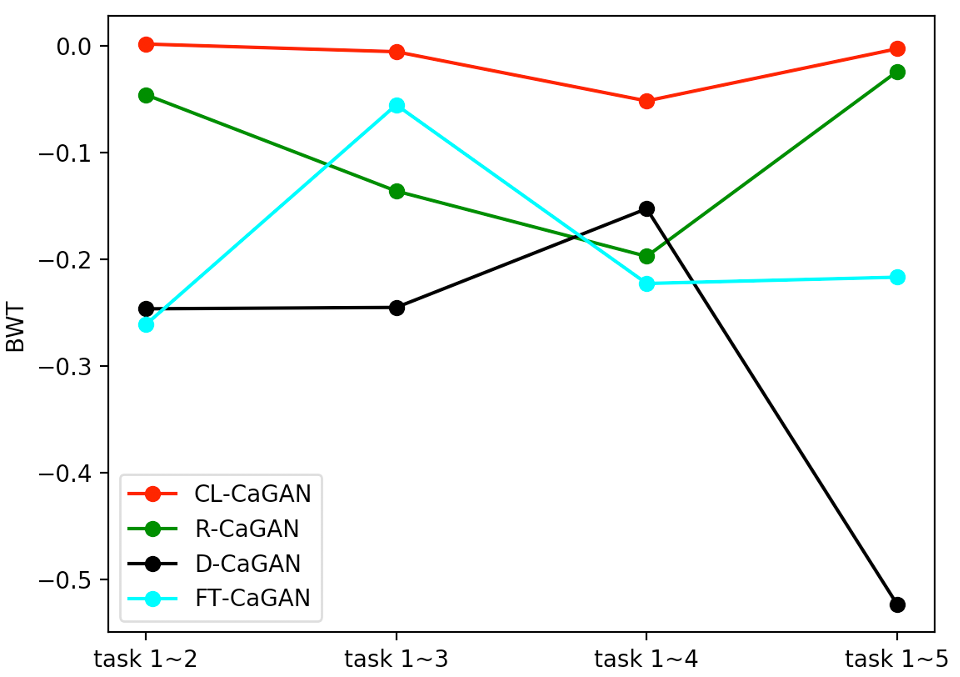}
\end{minipage}
}%插入图片，[]中设置图片大小，{}中是图片文件名

\caption{Performance obtained by different methods in open scenario. (a) AUC of each task. (b) The change of ACC as new task comes. (c) The change of BWT as new task comes.} %最终文档中希望显示的图片标题htb
\label{Rec} %用于文内引用的标签
\end{figure*}

The quality of reconstruction is evaluated by the reconstruction error between the input spectral vector and the reconstructed pseudo vector. Fig. \ref{RecBA} displays the difference of original and reconstructed spectral vectors on anomalous pixels and background pixels from Cat Island data sets. We can observed that anomalies in HSIs are not well recovered, whereas background pixels are reconstructed by CaGAN with less reconstruction error.

(iii) Effect Analysis of Threshold $\beta$ for CBM

In equation (\ref{eq6}), the threshold $\beta$ plays an important role in selection background samples in CBM procedure. A large number of anomaly samples will be selected into the background sample set $\textbf{B}$ when setting a small threshold value and resulting in contaminated and impure background information of the model learning. Conversely, the samples extremely similar to their neighbors are involved without considering diversity of the background samples located in different area
while the threshold is set too large. The effect of different thresholds on the detection performance of CaGAN is analyzed in Fig.\ref{threshold}. We can observe that the model achieves the optimal detection accuracy at a threshold value of 0.99 for all different datasets.

\subsection{Cross-Domain Detection Performance of CL-CaGAN}

In this part, we mainly analyze and discuss CL-CaGAN  through the performance of different scenes of HAD in open scenario circumstance. The implementation details, evaluation metrics and detection performance in open scenario are illustrated in the following part.

%1. Implementation of the CL-CaGAN in open scenario
\subsubsection{Comparison Methods of CL}
In the real application, the HAD task usually coming with an unending cross-scene detection tasks, while traditional DL cannot adaptive to different spectral dimension of HSIs due to the model hyperparameter can not change with different scenarios. Considering traditional DL-based algorithm cannot deal with open scenario circumstance, the results are usually incapable of adapting to the previous tasks and further tasks, which can present satisfying result for the current task. Therefore, we exploit and provide a new CL-CaGAN structure with new deliberated loss and replay mechanism to mitigate catastrophic forgetting problem caused in open scenario HAD tasks. Meanwhile, this research work is the first work for dealing with the catastrophic forgetting problem in HAD task. Therefore, in order to demonstrate the effect of our proposed CL-CaGAN structure, we compare the CL-CaGAN with Fine tune-based CaGAN(FT-CaGAN), Distillation-based CaGAN(D-CaGAN), Replay-based CaGAN(R-CaGAN) and Joint learning-based CaGAN(J-CaGAN) on several cross-scene HAD tasks to verify the robustness and advantageous of our proposed algorithm. To cope with the diversity and difference of spectral dimension in different scenarios, we adopt principal components analysis (PCA) \cite{kang2017pca} to unify the dimension of input HSIs, which achieves a universal model for dealing with varied spectral dimensions of all the previous tasks and the future task in one unified training process.

\subsubsection{Evaluation Metrics for CL}

We adopt average accuracy (ACC) as the average AUC of all tasks. To measure the capacity of remembering, backward transfer (BWT) is reported to evaluate how much new tasks influence the performance on previous tasks. The larger BWT score represents better performance for alleviating catastrophic forgetting phenomena and indicates how new tasks help with the preceding tasks. The calculation formula of ACC and BWT can be rewritten as
\begin{equation}
\begin{split}
ACC=\frac{1}{T}\sum_{i=1}^{T}{AUC}_{T,i}
\end{split}
\end{equation}
\begin{equation}
\begin{split}
BWT=\frac{1}{T-1}\sum_{i=1}^{T-1}{{AUC}_{T,i}-{AUC}_{i,i}}
\end{split}
\end{equation}
where ${AUC}_{T,i}$ is the test classification accuracy on task $i$ after sequentially learning unending $t$-th task.

\begin{table}[ht]
	\centering
	\caption{ Analyzing the impact of the number of clustering groups on CL}
	\label{analysis_K}
	\setlength{\tabcolsep}{2.5 mm}{
	\begin{tabular}{cccccc}
		\toprule  % 顶部线
		\multicolumn{2}{c}{k} & 1-2 Tasks	& 1-3 Tasks & 1-4 Tasks & 1-5 Tasks \\
        \midrule  % 中部线
        \multirow{2}*{k=2} & ACC & \textbf{0.9766} & 0.8276 & 0.7216 &  0.8965\\
        ~ & BWT & -0.0021 & -0.2362 & -0.3356 &  -0.0993\\
        \midrule  % 中部线
		\multirow{2}*{k=3} & ACC & \textbf{0.9766} & \textbf{0.9797} & \textbf{0.9153} & \textbf{ 0.9577}\\
        ~ & BWT & \textbf{0.0013} & \textbf{-0.0058} & \textbf{-0.0519} &  \textbf{-0.003}\\
        \midrule  % 中部线
        \multirow{2}*{k=4} & ACC & 0.9695 & 0.9294 & 0.5698 &  0.7151\\
        ~ & BWT & \textbf{0.0013} & -0.0058 & -0.4103 &  -0.2258\\
        \midrule  % 中部线
        \multirow{2}*{k=5} & ACC & 0.9693 & 0.7482 & 0.7417 & 0.8243\\
        ~ & BWT & 0.0008 & -0.3379 & -0.2759 & -0.1636\\
        \midrule  % 中部线
        \multirow{2}*{k=6} & ACC & 0.8714 & 0.8583 & 0.5275 & 0.6187\\
        ~ & BWT & -0.1371 & -0.1406 & -0.5421 & -0.4060\\
        
		\bottomrule  % 底部线
	\end{tabular}
	}
	
\end{table}

In Table \ref{analysis_K}, we first analyze the effect of the number of clustering groups on CL. It is shown that when the data are clustered into 2-6 groups respectively, the complexity and representative of the information contained in the replay buffer are also different. When the number of groups for clustering is 2, it is obvious that the replayed samples can not be well represented, resulting in catastrophic historical feature forgetting problem. As the increased clustering groups, more and more complex features are retained, which will bring more disruption for the modal learning of the new coming tasks. Therefore, from above comparison, we select the most appropriate clustering hyperparameters as 3 in the following experiments.

The ACC and BWT scores of different number of tasks are presented in Table \ref{ACCBWT} (The optimal scores are in bold). Five datasets (tasks) are coming sequentially: (1) Los Angeles-1, (2) Los Angeles-2, (3) Bay Champagne, (4) San Diego, (5) Cat Island. For more comprehensively comparison, we evaluate each comparison algorithm by ACC and BWT scores when each new task arrives, where $t$-th task in Table \ref{ACCBWT} represents that there are already $t$ datasets ($t$ tasks) have been trained by now. The CaGAN and J-CaGAN train the datasets separately, which are incompetent to catastrophic forgetting. Among all algorithm, the proposed CL-CaGAN achieves more stable ACC and BWT, which reveals that our method better balances the performance of previous tasks and current task. 

After training the network for various tasks in open scenario, the model has the ability to detect anomalies for all previous tasks. The AUC of each previous task is presented in Fig.\ref{AUC}, which reveals that all methods present well on current 5-th task, but FT-CaGAN, J-CaGAN, D-CaGAN and CaGAN suffer from serious catastrophic forgetting on previous tasks. As new data comes, the change of ACC is shown in Fig.\ref{ACC}, which indicates that the proposed CL-CaGAN demonstrates the highest and the most robust performance with new coming datasets. Fig. \ref{BWT} illustrates the changes of BWT. The BWT of D-CaGAN degrades rapidly when the 5-th task comes, while the performance of R-CaGAN and FT-CaGAN deteriorate when the 4-th task comes. Whereas our proposed CL-CaGAN presents insensitive to catastrophic forgetting problem, which indicates less sensitivity for all the different anomaly detection scenarios. From above illustrated experimental results, we can further conclude that CL-CaGAN realizes a equilibrium between remembering of history knowledge and adaptation of new arrived tasks.

\section{Conclusion}

In this paper, a CL-CaGAN is proposed for improving detection performance and alleviating the catastrophic forgetting phenomenon in cross-domain HAD task. The continuous exemplar replay strategy with self-distillation loss is constructed for retaining history knowledge and adapting to the new arrived tasks in open scenario situation. Meanwhile, the proposed CL-CaGAN with differentiable data augmentation realizes an end-to-end reconstruction by cooperating a modified capsule structure in an elegant way as the generator and discriminator with GAN for effectively learning representative spectral characteristics of background distribution, and further ensures stability and equilibrium of the training procedure for the whole structure. Experiments on five real HAD datasets demonstrate that the proposed CL-CaGAN presents more satisfying capability for anomaly detection, and demonstrates more robust detection performance with considering a equilibrium between history tasks and new arrived tasks for cross-scene HAD, which paves a new way for practical application of DL structure in open scenario HAD circumstance.
%\section*{Acknowledgments}
%The authors would like to thank the handling editor and anonymous reviewers for their valuable comments and suggestions, which greatly improved the quality of this article.

\bibliographystyle{IEEEtran}
\bibliography{main}
\vspace{-10 mm}
\begin{IEEEbiography}[{\includegraphics[width=1in,height=1.25in,clip,keepaspectratio]{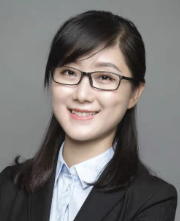}}]{Jianing Wang}
(Member, IEEE) received the B.S. and M.S. degrees in circuit and system from Lanzhou University, Lanzhou, China, in 2005 and 2008, respectively, and the Ph.D. degree from Xidian University, Xi’an, China, in 2016. 

She worked with China Aerospace Science and Technology Corporation, Xi’an. She is an Associate Professor and a Member of the Key Laboratory
of Intelligent Perception and Image Understanding, School of Computer Science and Technology, Ministry of Education of China, Xidian University, Xi’an. Her research interests include image processing, machine learning, and artificial intelligent algorithm and applications. Her research directions include big data processing, embedded algorithms for intelligent model compression methods, pattern recognition and artificial intelligence, video data processing, analysis and content understand and continual learning.
\end{IEEEbiography}

\begin{IEEEbiography}[{\includegraphics[width=1.1in,height=1.25in,clip,keepaspectratio]{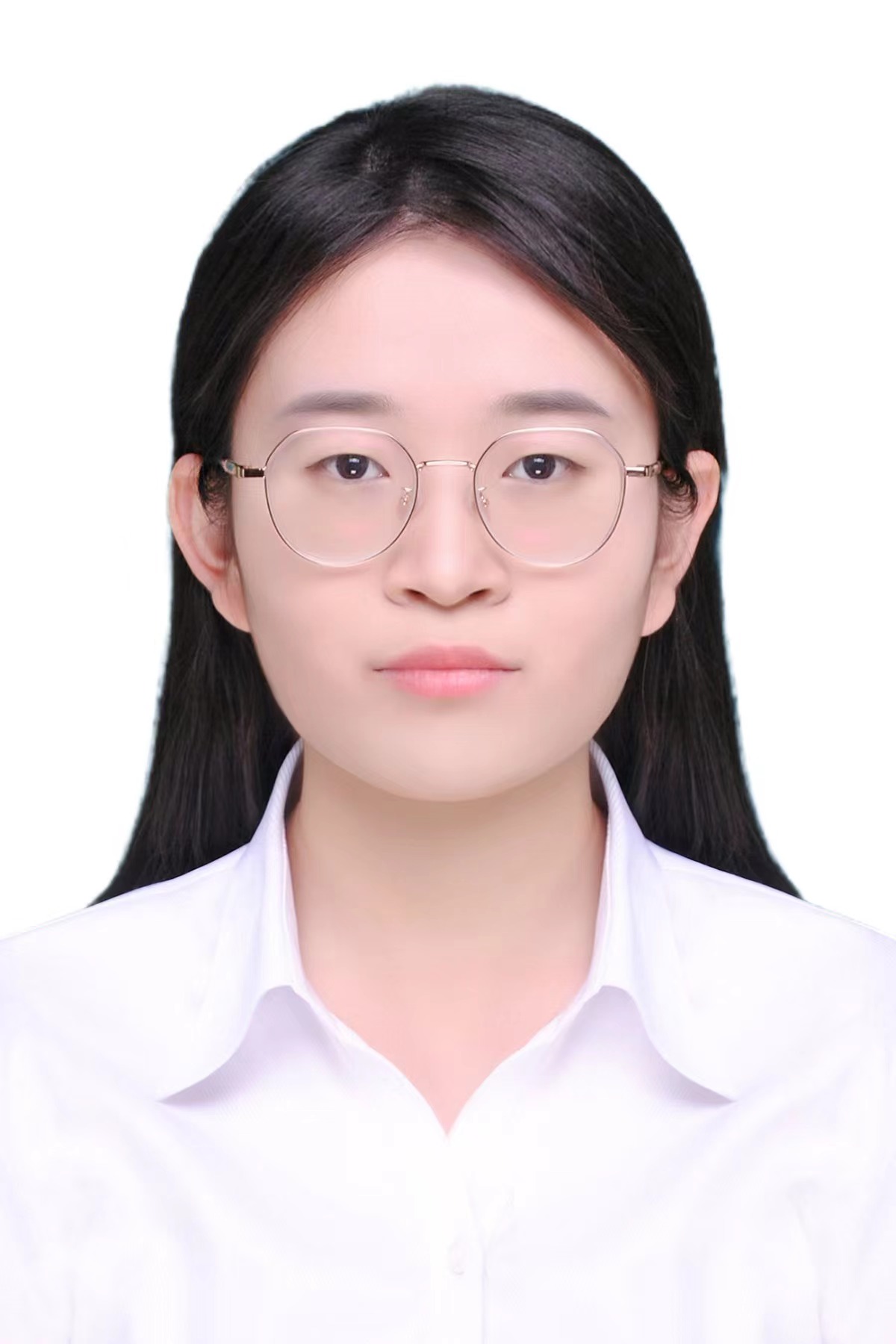}}]{Siying Guo} received the bachelor's degree from the School of Computer Science and Technology in 2020, and the master's degree from the School of Artificial Intelligence in 2023.
 
Her research interests include deep learning, remote sensing data processing, and image processing.\end{IEEEbiography}
\begin{IEEEbiography}[{\includegraphics[width=1in,height=1.25in,clip,keepaspectratio]{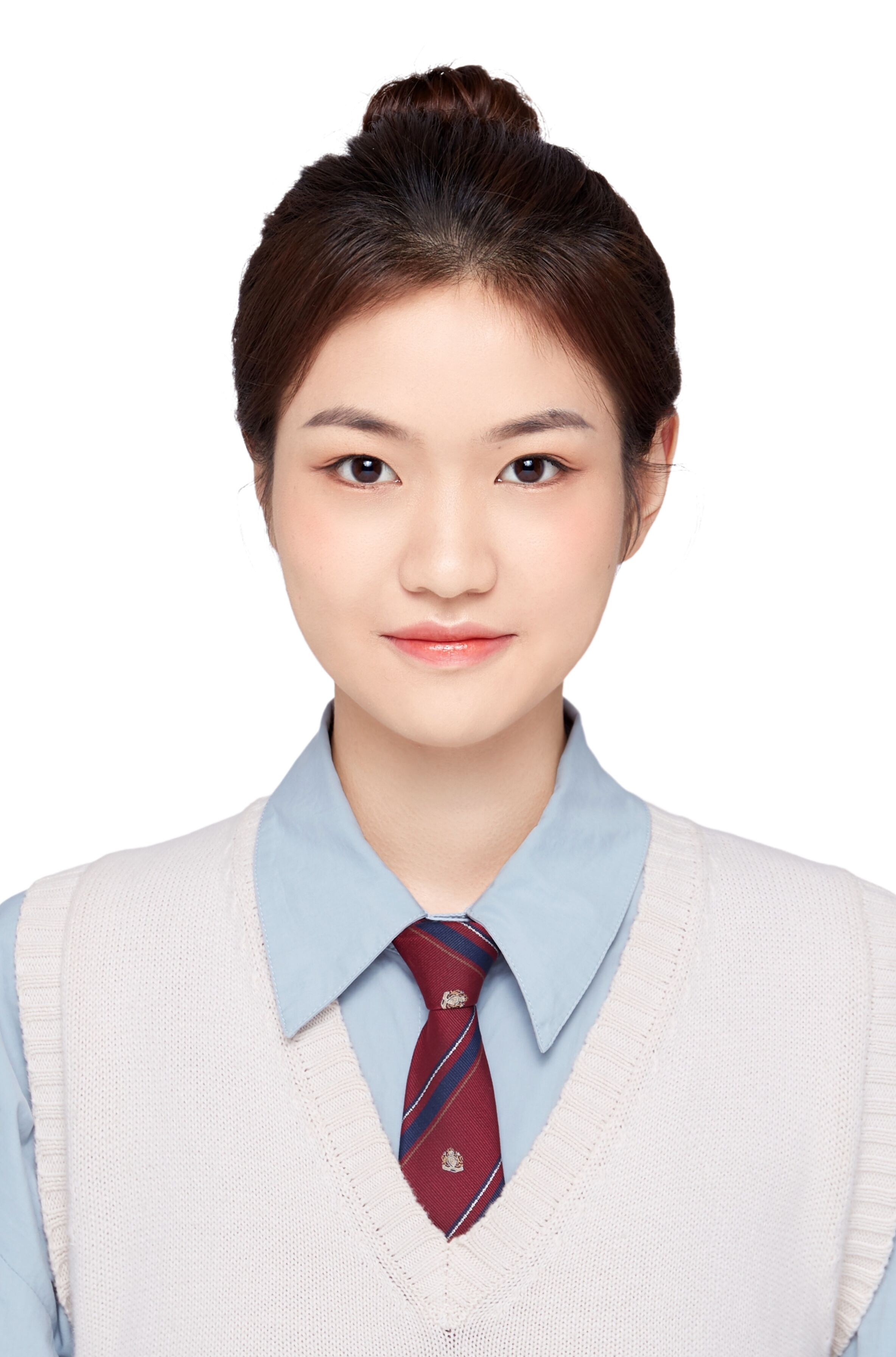}}]{Zheng Hua}
is currently pursuing a Master's degree in Computer Science and Technology, Xidian University, Xi’an, China.
 
Her research interests include continual learning, incremental learning, hyperspectral anomaly detection and image processing.
\end{IEEEbiography}
\vspace{-5 mm}
\begin{IEEEbiography}[{\includegraphics[width=1in,height=1.25in,clip,keepaspectratio]{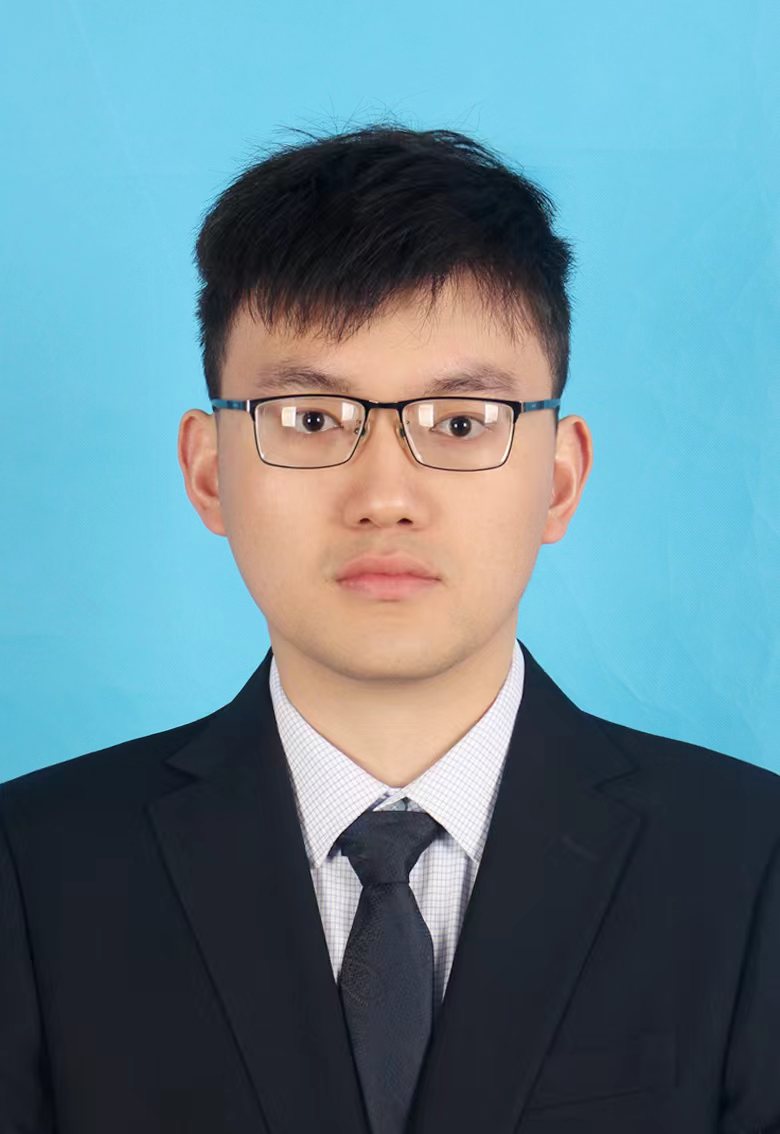}}]{Runhu Huang} received the bachelor’s degree from the School of Intelligence Science and Technology in 2020, Xidian University, Xi’an, China, in 2020, where he is pursuing the master’s degree with the School of Artificial Intelligence in 2023.

His research interests include model compression, deep learning, and remote sensing image processing.\end{IEEEbiography}
\vspace{-5 mm}
\begin{IEEEbiography}[{\includegraphics[width=1in,height=1.25in,clip,keepaspectratio]{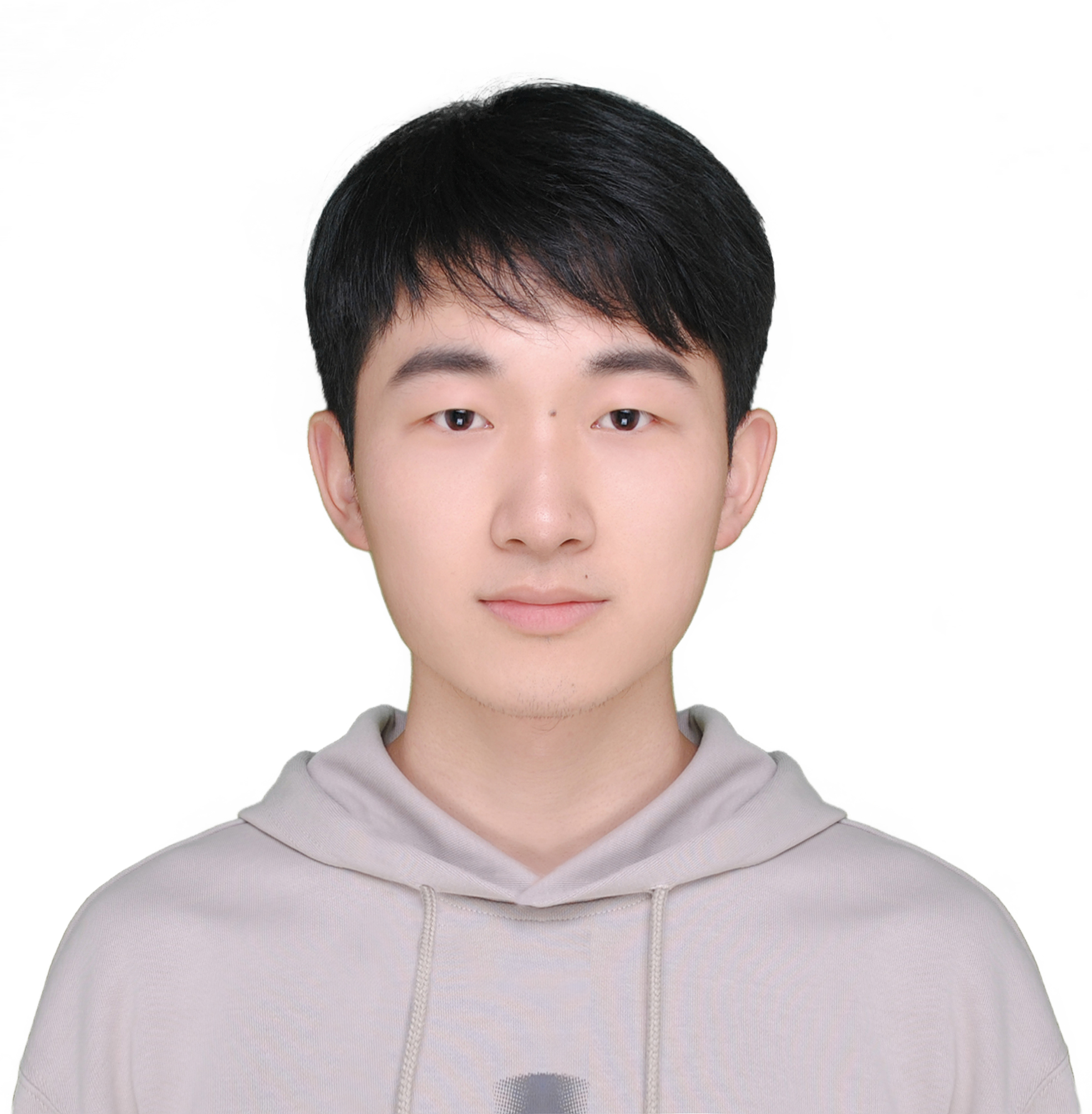}}]{Jinyu Hu} is currently pursuing a Master's degree in Computer Science and Technology, Xidian University, Xi’an, China. 

His research interests include model compression, remote sensing image processing, and neural architecture search.\end{IEEEbiography}
\vspace{-5 mm}
\begin{IEEEbiography}[{\includegraphics[width=1in,height=1.25in,clip,keepaspectratio]{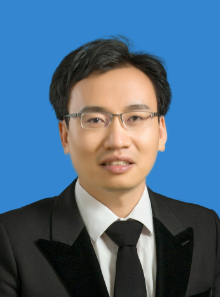}}]{Maoguo Gong} (M’07-SM’14-F’24) received the B.Eng. degree and Ph.D. degree from Xidian University. Since 2006, he has been a teacher of Xidian University. He was promoted to associate professor and full professor in 2008 and 2010, respectively, both with exceptive admission.

Gong’s research interests are broadly in the area of computational intelligence, with applications to optimization, learning, data mining and image understanding. He has published over one hundred papers
in journals and conferences, and holds over twenty granted patents as the first inventor. He is leading or has completed over twenty projects as the Principle Investigator, funded by the National Natural Science Foundation of China, the National Key Research and Development Program of China. He was the recipient of the prestigious National Program for Support of the Leading Innovative Talents from the Central Organization Department of China, the Leading Innovative Talent in the Science and Technology from the Ministry of Science and Technology of China, the Excellent Young Scientist Foundation from the National Natural Science Foundation of China, the New Century Excellent Talent from the Ministry of Education of China, and the National Natural Science Award of China.
 \end{IEEEbiography}

\end{document}